\documentclass[lettersize,journal]{IEEEtran}

\IEEEoverridecommandlockouts                              

\usepackage[pdftex]{graphicx}
\usepackage{hhline}
\usepackage{amsmath}
\usepackage{amssymb}
\usepackage{booktabs, makecell, multirow, rotating}
\usepackage{graphicx}
\usepackage{float}
\usepackage{caption}
\usepackage{cite}
\usepackage{textcomp}
\usepackage{subfigure}
\usepackage{multicol, blindtext}
\usepackage{tabularx}
\usepackage{hyperref}
\usepackage[dvipsnames]{xcolor}
\usepackage{enumitem}
\usepackage{gensymb}
\usepackage{algorithm}
\usepackage{algorithmic}
\usepackage{microtype} 
\usepackage{colortbl}
\usepackage{xcolor}
\usepackage{array}
\newcolumntype{C}[1]{>{\centering\arraybackslash}m{#1}}

\title{\LARGE \bf
Learning Bimanual Cloth Manipulation with Vision-based Tactile Sensing via Single Robotic Arm
}

\author{Dongmyoung~Lee,~Wei~Chen,~Xiaoshuai~Chen,~Rui~Zong,~and~Petar~Kormushev,~\IEEEmembership{Member,~IEEE}
\thanks{Dongmyoung Lee, Wei Chen, Xiaoshuai Chen, Rui Zong, and Petar Kormushev are with the Robot Intelligence Lab, Dyson School of Design Engineering, Imperial College London, 25 Exhibition Road, London, SW7 2DB, UK
{\tt\footnotesize (d.lee20, w.chen21, c.xiaoshuai19, rui.zong21, p.kormushev)@imperial.ac.uk}}
}

\begin{document}

\maketitle





\begin{abstract}
Robotic cloth manipulation remains challenging due to the high-dimensional state space of fabrics, their deformable nature, and frequent occlusions that limit vision-based sensing. Although dual-arm systems can mitigate some of these issues, they increase hardware and control complexity. This paper presents \emph{Touch G.O.G.}, a compact vision-based tactile gripper and perception/control framework for single-arm bimanual cloth manipulation. The proposed framework combines three key components: (1) a novel gripper design and control strategy for in-gripper cloth sliding with a single robot arm, (2) a \emph{Vision Foundation Model}-backboned Vision Transformer pipeline for cloth part classification (\emph{PC-Net}) and edge pose estimation (\emph{PE-Net}) using real and synthetic tactile images, and (3) an encoder-decoder synthetic data generator (\emph{SD-Net}) that reduces manual annotation by producing high-fidelity tactile images. Experiments show \emph{96\%} accuracy in distinguishing edges, corners, interior regions, and grasp failures, together with sub-millimeter edge localization and $\mathbf{4.5^\circ}$ orientation error. Real-world results demonstrate reliable cloth unfolding, even for crumpled fabrics, using only a single robotic arm. These results highlight \emph{Touch G.O.G.} as a compact and cost-effective solution for deformable object manipulation. Supplementary material is available at https://sites.google.com/view/touchgog.
\end{abstract}


\section{Introduction}
The robotic manipulation of deformable objects, particularly fabrics, represents a frontier in mechatronics where traditional rigid-body control strategies often fail. The high-dimensional state space and unpredictable dynamics of cloth require systems that can adapt in real time to local deformations. While dual-arm robotic systems have shown success in tasks such as unfolding, they introduce substantial hardware cost and control complexity, which limits their deployment in unstructured domestic and industrial environments~\cite{borras2022effective}. Another major bottleneck in cloth manipulation is perception under occlusion. Traditional vision-based approaches rely on global cameras to identify features such as corners~\cite{jimenez2020perception,10342086}. However, during intricate tasks such as edge tracing or hand-over-hand sliding, the robot end-effector and fabric folds frequently occlude the camera view, often resulting in open-loop failure.

\begin{figure}[t!]
    \centering
    \includegraphics[width=0.98\columnwidth]{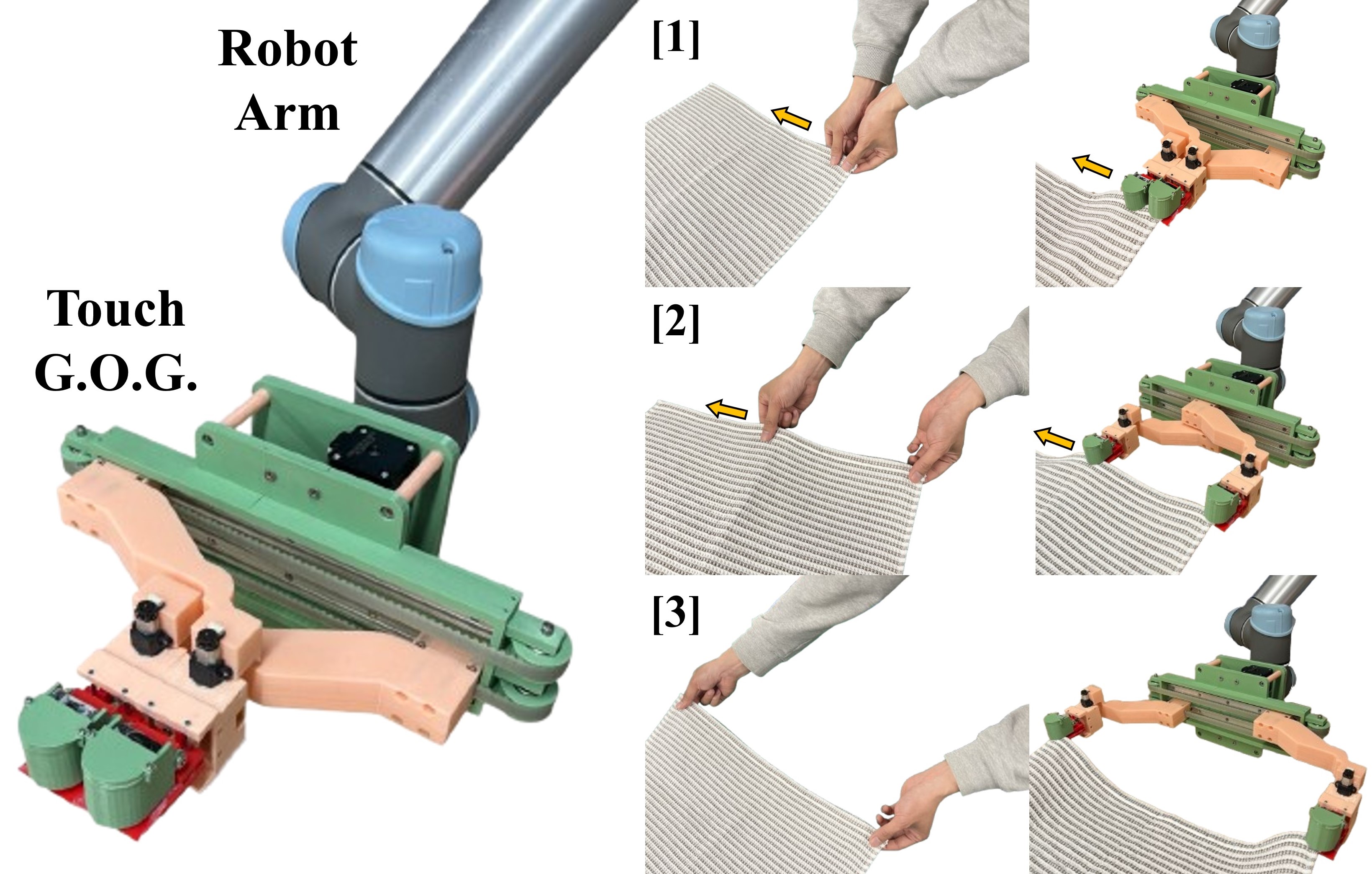}
    \caption{\emph{Touch G.O.G.} framework enabling single-arm bimanual cloth manipulation. Unlike rigid grippers, our system utilizes a human-inspired sliding strategy: (1) The robot identifies structural regions (e.g., corners), (2) The gripper actively expands and modulates friction to slide along the edge, and (3) Real-time tactile feedback corrects pose errors during sliding until the opposing corner is reached.}
    \label{fig:concept_overview}
\end{figure}

To overcome this, we propose a paradigm shift from global vision to active visuotactile local control. We introduce \emph{Touch G.O.G.}, a mechatronic system designed to emulate bimanual dexterity using a single arm. Unlike previous mechanical solutions that relied on passive compliance or open-loop kinematics~\cite{lee2024gog}, our approach integrates high-resolution tactile sensing directly into the control loop. This allows the system to perform closed-loop edge tracing and error correction that was previously impossible with mechanical constraints alone. However, robust learning-based tactile perception is fundamentally data-limited. Accurate supervision for edge geometry in tactile images is expensive to collect and difficult to annotate at scale across diverse fabrics. To address this data bottleneck, we introduce SD-Net, a SAM-backboned encoder-decoder framework that generates high-fidelity synthetic tactile images from lightweight edge annotations, enabling data-efficient training of PE-Net.

This paper presents a unified robotic system comprising three contribution:

\noindent \begin{enumerate}
    \item \textbf{Novel Gripper Design and Control Strategy:} We introduce a visuotactile end-effector and control strategy for single-arm bimanual cloth manipulation. The proposed design combines decoupled belt-driven width control with an additional abduction degree-of-freedom (DoF), enabling adaptive grasp modulation, in-gripper cloth sliding, and real-time pose correction during manipulation.
    
    \item \textbf{Foundation-Model-Based Visuotactile Perception (\emph{PC-Net} and \emph{PE-Net}):} We present a robust perception pipeline built on the Segment Anything Model (SAM)~\cite{Kirillov_2023_ICCV}, comprising a cloth part classification network (\emph{PC-Net}) and an edge pose estimation network (\emph{PE-Net}). Together, these networks enable reliable recognition of grasp states and precise edge localization, achieving sub-millimeter position accuracy even on complex, patterned fabrics.
    
    \item \textbf{Synthetic Data Generator (\emph{SD-Net}):} To address the limited availability of annotated tactile datasets, we propose a SAM-backboned encoder-decoder framework that synthesizes high-fidelity tactile images from edge annotations, significantly reducing manual labeling while expanding pose and texture diversity for training \emph{PE-Net}.
\end{enumerate}

We validate this framework through extensive real-world experiments, demonstrating reliable unfolding of crumpled fabrics using a single robotic arm which requires the synergy of mechanical dexterity and algorithmic intelligence.

\begin{figure*}[h]
    \centering
    \includegraphics[width=0.98\linewidth]{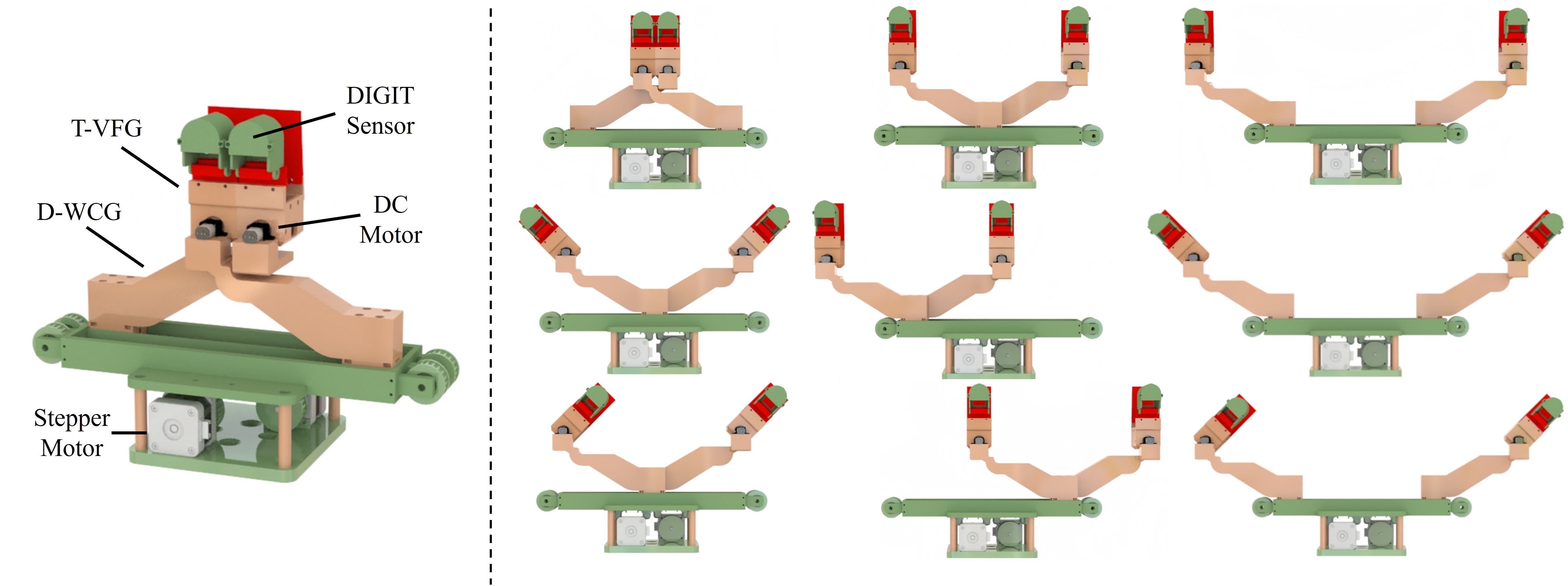}
    \caption{\textit{Touch G.O.G.} overview: (left) the CAD model of the \textit{Touch G.O.G.}, showing the base, the Decoupled Width Control Gripper (D-WCG), and the two Tactile Variable Friction Grippers (T-VFGs) and (right) the versatility of D-WCG and T-VFG.}
    \label{fig:tgog_system}
\end{figure*}

\section{Related works} \label{sec:relatedworks}
\subsection{Cloth Unfolding}
Cloth unfolding is a fundamental process for downstream tasks like ironing, folding, and dressing. Existing research can be broadly grouped into quasi-static and dynamic manipulations. Quasi-static methods focus on sequential pick-and-place operations to unfold cloth. These approaches generally rely on identifying cloth features such as wrinkles, edges, or corners~\cite{martinez2018recognition}. Learning-based methods have been employed to learn optimal actions for cloth unfolding~\cite{ebert2018visual, saxena2019garment, seita2019deep, wu2019learning, hietala2022learning}. In addition to this, other approaches also aim to explicitly learn the dynamics of the garment model by Graph Neural Network (GNN)-based approaches~\cite{chen2025graphgarment}. These strategies typically require many actions and degrade in performance when key features are heavily occluded. Dynamic manipulation aims to achieve rapid cloth coverage with a few non-prehensile actions~\cite{ha2022flingbot, jangir2020dynamic, xu2022dextairity}. While effective, these methods typically require elaborate setups involving multiple robotic arms or external air control systems, increasing both system cost and complexity.

An alternative strategy is \emph{edge tracing}, where the gripper slides along a cloth edge~\cite{yuba2017unfolding, gabas2021dual, garcia2020benchmarking}. Recent advancements have enhanced edge sliding by incorporating tactile sensing, emulating human dexterity in handling deformable objects. Vision-based tactile systems have demonstrated the ability to trace cloth edges and detect corners for unfolding tasks~\cite{sunil2023visuotactile, zhang2023visual}. However, these methods are generally demonstrated on simple fabrics and struggle with patterned or textured cloth, where pose estimation and feature detection become more complex. In this study, we build upon these approaches by leveraging vision-based tactile sensing to enhance edge sliding performance. Our method addresses challenges associated with diverse cloth by integrating advanced perception and control strategies.

\subsection{Data Augmentation}
Data augmentation is a common strategy to alleviate data scarcity by artificially expanding training distributions and improving generalization~\cite{perez2017effectiveness}. Classical techniques, such as affine transforms and color jittering~\cite{bjerrum2017smiles}, are widely used, but aggressive augmentation can distort the underlying data distribution and introduce biases~\cite{perez2017effectiveness}.

To go beyond these transformations, generative models have been explored for data augmentation. GAN-based methods~\cite{gulrajani2017improved, ZHENG20201009, Karras2021} can produce high-fidelity, photorealistic images. Hybrid GAN-diffusion approaches further increase sample diversity for complex patterns~\cite{wang2023diffusiongan}, but typically at the cost of substantial computational overhead and non-trivial training stability issues. Diffusion models have recently emerged as a powerful alternative for data augmentation~\cite{azizi2023synthetic, li2023synthetic}. Variants such as the Denoising Diffusion Implicit Model (DDIM) improve sampling efficiency while preserving generation quality~\cite{song2021denoising}, yet most of pipelines still incur substantial computational cost due to the large number of denoising steps required during sampling~\cite{dhariwal2021diffusion}.

In robotics, self-annotated scene pipelines can generate pixel-wise masks in cluttered environments and have successfully trained segmentation networks for manipulation~\cite{8460950} and cluttered grasping~\cite{10532168}. Our work follows this line but targets visuotactile sensing. We propose a SAM-backboned encoder–decoder framework (\emph{SD-Net}) that generates high-fidelity tactile images directly from edge annotations, enabling to train robust cloth edge pose estimation with minimal manual labeling.

\subsection{Vision-based Tactile Manipulation}

Vision-based tactile sensors (GelSight~\cite{yuan2017gelsight} and DIGIT~\cite{lambeta2020digit}) have emerged as powerful tactile sensors by offering high spatial resolution and fast response times. These sensors have been successfully employed system property identification, such as cloth texture recognition~\cite{yuan2018active} and liquid classification~\cite{huang2022understanding}. In manipulation tasks, vision-based tactile sensing has demonstrated significant potential by enabling robust control. For example, they have been utilized for in-hand ball-rolling manipulation using deep tactile model predictive control~\cite{tian2019manipulation, lambeta2020digit}, enhancing grasping stability~\cite{kolamuri2021improving}, precise slip detection~\cite{taylor2022gelslim}, and accurate object surface following~\cite{lin2022tactile}. 

Recent works has extended vision-based tactile sensing to deformable object manipulation~\cite{tirumala2022learning, she2021cable, pecyna2022visual}. Specifically within the domain of cloth manipulation, vision-based tactile sensing has been used for cloth edge tracing with a Linear–Quadratic Regulator (LQR) controller~\cite{sunil2023visuotactile} or offline RL policies~\cite{zhou2021plas}. However, most of these approaches are limited to manipulating small square cloths and often require dual-arm robotic setups, restricting flexibility and practicality in single arm-based bimanual tasks.

In this paper, we present the \textit{Touch G.O.G.}, a novel system that integrates vision-based tactile sensing with a SAM-backboned Vision Transformer (ViT) and an encoder-decoder-based synthetic dataset pipeline. This cohesive framework significantly improves edge detection, pose estimation, and controlled sliding by addressing challenges related to occlusions, complex patterns, and diverse cloth configurations. Our system achieves robust single-arm manipulation, leveraging the synergy between advanced perception and mechanical innovation.

\section{Touch G.O.G. System} \label{sec:methodology}
This section details the mechanical design of the \textit{Touch G.O.G.} as depicted in Fig.~\ref{fig:tgog_system}. To enable closed-loop tactile sliding, the system integrates two functionally distinct modules: Decoupled Width Control Gripper (D-WCG) for global span regulation and Tactile Variable Friction Gripper (T-VFG) for local grasp modulation and sensing.

\subsection{Decoupled Width Control Gripper (D-WCG)}\label{sec:decoupled_wcg}
D-WCG serves as the prismatic base of the system, providing the translational degrees of freedom necessary to emulate bimanual spreading and tensioning. The mechanism utilizes a dual-rail linear stage where each finger carriage is independently actuated by a dedicated stepper motor via a timing belt transmission. This decoupled actuation strategy enables asymmetric finger positioning, allowing the system to adapt its grasp width dynamically to irregular fabric geometries and varying cloth sizes (Fig.~\ref{fig:tgog_system}). To ensure precise control of the tensioning forces applied to the fabric, a belt-driven transmission is selected for its high stiffness and minimal backlash. Unlike compliant or underactuated mechanisms, this rigid transmission ensures that the linear separation of the fingers tracks the motor position commands accurately, independent of the resistive loads imposed by the stretching fabric. This direct electromechanical coupling is essential for performing stable, closed-loop expansion maneuvers during cloth unfolding.

\subsection{Tactile Variable Friction Gripper (T-VFG)}\label{sec:tactile_vfg}

\begin{figure}[t]
    \centering
    \includegraphics[width=0.98\columnwidth]{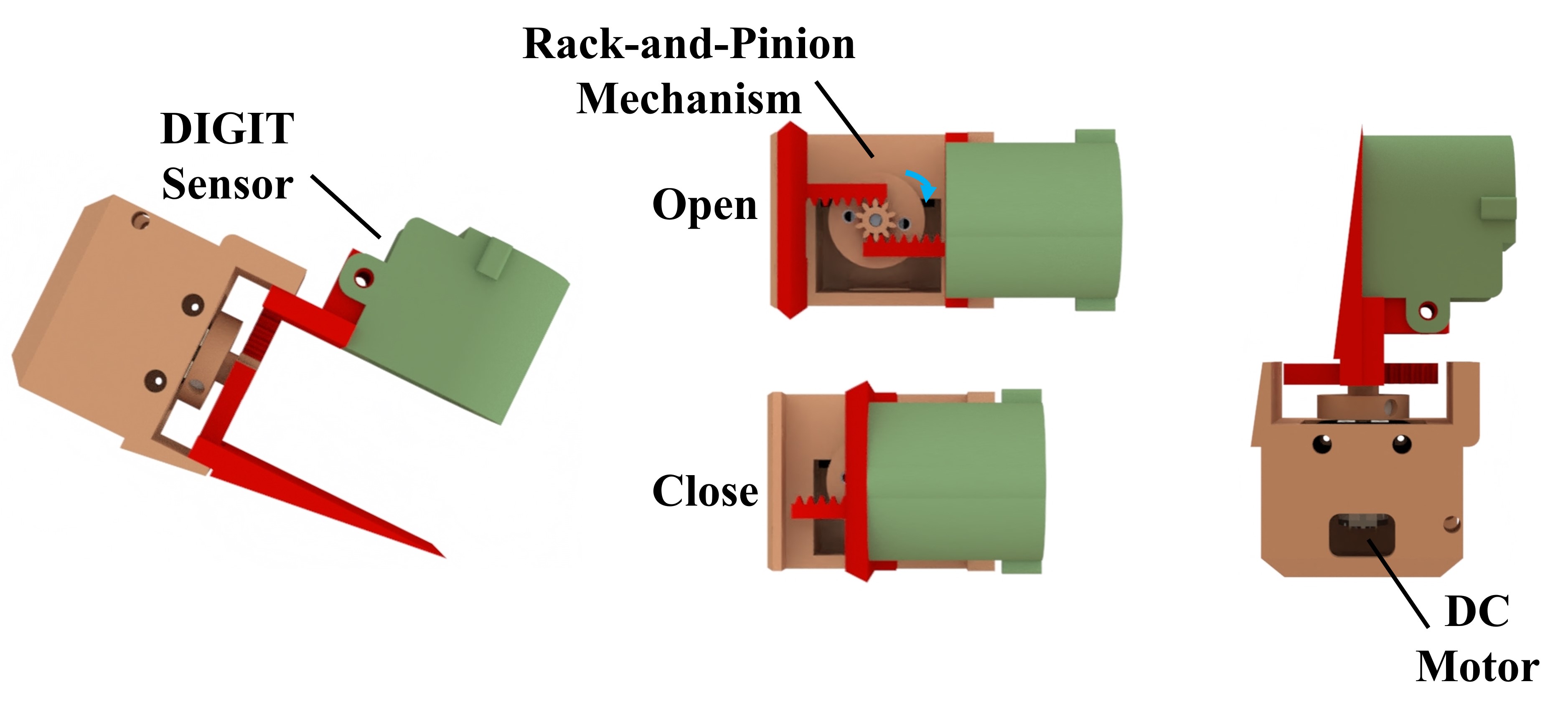}
    \caption[Detail of the T-VFG in the \textit{Touch G.O.G.} system.]{Detail of the T-VFG in the \textit{Touch G.O.G.} system, showing the DIGIT sensor for high-resolution tactile feedback and the DC motor that drives the rack-and-pinion mechanism.}
    \label{fig:tgog_vfg}
\end{figure}

T-VFG module (Fig.~\ref{fig:tgog_vfg}) acts as the active interaction interface of the system. This design targets three components that are critical for reliable in-gripper sliding under occlusion: (i) sufficient kinematic redundancy to re-orient the contact, (ii) high-resolution contact sensing to resolve local fabric features, and (iii) closed-loop control for deformable objects.

Each T-VFG is mounted at the distal end of a D-WCG finger, forming a finger with an additional abduction DoF. The abduction axis lies approximately in the plane orthogonal to the primary grasping direction and is actuated by a compact DC motor housed inside the D-WCG finger, transfering torque directly to the rotating T-VFG. Mechanically, this abduction DoF is actuated via a small DC motor housed within the D-WCG finger body, which transfers torque directly to the rotating T-VFG. 

The \textit{Touch G.O.G.} incorporates a vision-based tactile sensor (DIGIT) on each finger pad. The DIGIT sensor consists of an internal camera and an elastomeric sensing surface to capture high-resolution images of the contact interface, providing a dense measurement of local surface geometry such as corners, edges, regions inside the fabric, or grasp failures. 

This capability allows the system to distinguish between successfully grasping a fabric layer and failing to make contact, preventing grasp failures. This tactile-driven adaptability significantly enhances the precision and robustness of cloth manipulation, particularly in tasks requiring delicate handling or real-time correction.

\subsection{T-VFG Closed-Loop Control}
Alongside the vision-based sensors, each T-VFG is driven by a DC motor equipped with an incremental encoder, enabling closed-loop position control of the grasping as well as the abduction angle. Both motions are governed by a standard PID controller that regulates the motor positions. Let \(e_k\) denote the error at time step \(k\), defined as the difference between the measured position \(\text{pos}_k\) and the desired setpoint \(\text{pos}_{t, k}\). The corresponding control command \(u_k\) is then calculated as:

{\footnotesize
\begin{equation}
  u_k \;=\; K_p\,e_k 
          \;+\; K_i \sum_{i=0}^{k} e_i\,\Delta T 
          \;+\; K_d\,\frac{e_k - e_{k-1}}{\Delta T},
\label{eq:pid_basics}
\end{equation}
}

\noindent
where \(K_p\), \(K_i\), and \(K_d\) are the proportional, integral, and derivative gains, respectively, and \(\Delta T\) is the sampling interval between two consecutive control updates.

Direct differentiation of encoder feedback can amplify measurement noise and introduce chattering in the control signal. To mitigate this effect, the derivation term is filtered using exponential smoothing:

{\footnotesize
\begin{equation}
  \dot{e}_{\text{filtered}, k} \;=\; (1 - \alpha)\,\dot{e}_{\text{filtered}, k-1} 
  \;+\; \alpha\,\biggl(\tfrac{e_k - e_{k-1}}{\Delta T}\biggr),
\label{eq:filtered_derivative}
\end{equation}
}

\noindent
where \(0 < \alpha < 1\) controls the trade-off between responsiveness and noise attenuation. The final PID control law becomes:

{\footnotesize
\begin{equation}
  u_k \;=\; K_p\,e_k \;+\; K_d\,\dot{e}_{\text{filtered}, k} 
  \;+\; K_i \sum_{i=0}^{k} e_i\,\Delta T.
\label{eq:pid_filtered}
\end{equation}
}

For the grasping motion, the controller precisely position the T-VFG fingers and the resulting motor effort is implicitly limited to avoid excessive normal force on the fabric. The abduction motion controls the rotational movement of the T-VFG, allowing the system to re-orient and adjust the gripper when it is misaligned or positioned too deeply in the fabric. The tuning parameters \(K_p, K_i, K_d\) and the smoothing factor \(\alpha\) are adjusted to account for the different dynamics and load conditions.

The control loop runs with a dynamically measured sampling interval $\Delta T$ to accommodate variations in loop timing typical of interrupt-driven embedded implementations. This explicit use of the measured $\Delta T$ preserves controller performance even when the execution period is not perfectly constant. Combined with the derivative filtering in Eq.~\ref{eq:filtered_derivative}, the overall scheme yields smooth motor responses with reduced oscillations, which is crucial for ensuring stable, repeatable sliding trajectories. By applying this controller to both the grasping and abduction motions, each T-VFG converges reliably to its commanded posture, enabling the \textit{Touch G.O.G.} to realize accurate in-hand adjustments, controlled slip, and compliant alignment with the garment edge.

\begin{figure}[t]
    \centering
    \includegraphics[width=0.98\columnwidth]{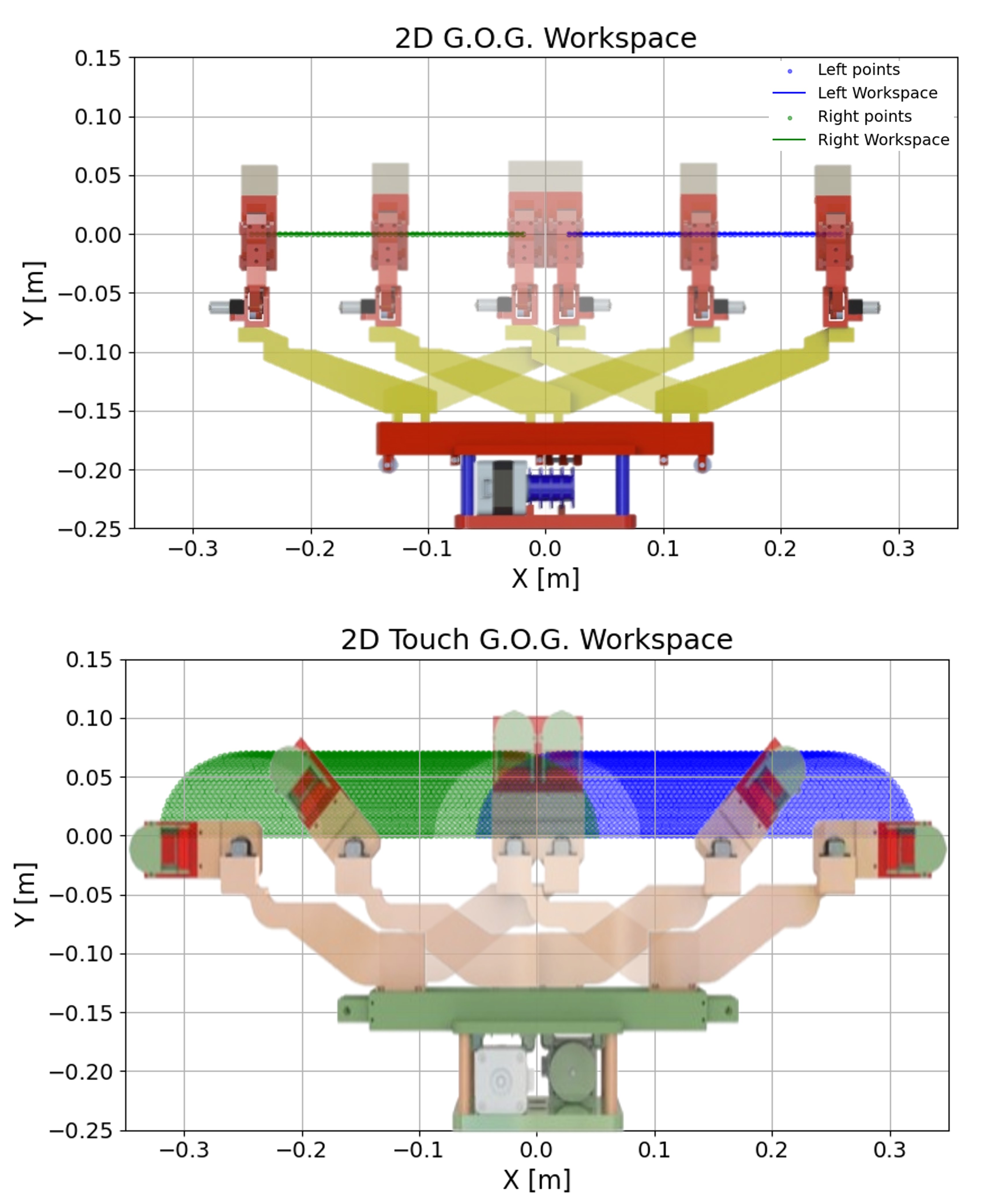}
    \caption[Comparison of workspaces between \textit{G.O.G.} and \textit{Touch G.O.G.}]{Comparison of workspaces: \textit{G.O.G.} (top) shows the left finger (blue) and right finger (green) workspaces, exhibiting lower dexterity. \textit{Touch G.O.G.} (bottom) shows larger coverage due to extra degrees of freedom, with left (green) and right (blue) regions superimposed on the CAD model.}
    \label{fig:gog_workspace}
\end{figure}

\begin{figure*}[h]
    \centering
    \includegraphics[width=0.98\linewidth]{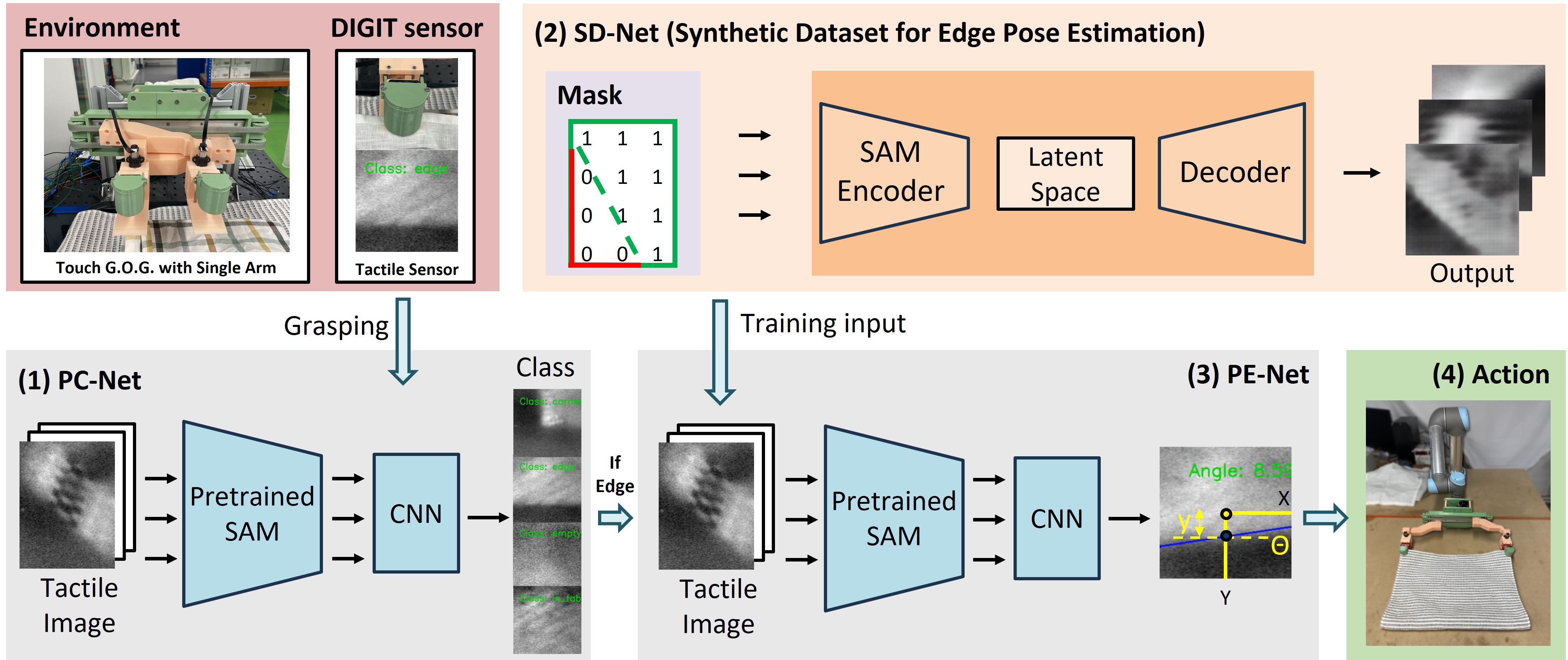} 
    \caption[Overview of the cloth manipulation framework with the use of single robotic arm.]{Overview of the cloth manipulation framework: The environment setup shows the \textit{Touch G.O.G.} system integrated with visuotactile sensors. (1) Cloth Part Classification Network (\textit{PC-Net}) is trained with a SAM backbone and a convolutional head. (2) The SAM-backboned Mask Decoder Network (\textit{SD-Net}) generates realistic synthetic tactile images from edge annotations. (3) The resulting synthetic dataset is used to train the Edge Pose Estimation Network (\textit{PE-Net}), which estimates edge pose for (4) precise cloth sliding.}
    \label{fig:gunfolding_description}
\end{figure*}

\subsection{Comparative Workspace Analysis}
Fig.~\ref{fig:gog_workspace} illustrate the planar workspaces associated with the \textit{Touch G.O.G.} system. Due to the decoupled linear motion and the additional abduction DoF on each finger, the proposed design generates a 2D reachable region in the plane. Within this region, each finger can be positioned and oriented continuously, enabling the end-effectors to approach cloth edges from multiple angles while maintaining contact in a single horizontal plane.

For reference, Fig.~\ref{fig:gog_workspace} also shows the workspace of a prior gripper configuration, whose fingers are constrained to a nearly 1D locus along the lateral axis. Although both mechanisms achieves a comparable maximum span along the $x$-direction, the abduction motion in \textit{Touch G.O.G.} substantially enlarges the set of reachable region. In practice, this broader and more flexible workspace allows the system to grasp objects that are offset from the central line and to adjust the approach angle to accommodate different cloth sizes, shapes, and edge orientations, which is essential for reliable single-arm bimanual manipulation.

\section{Vision-based Tactile Perception and Control} \label{sec:cpcnepen}
This section presents the visuotactile perception pipeline that underpins the closed-loop control behavior of \textit{Touch G.O.G.} system. As summarized in Fig.~\ref{fig:gunfolding_description}, the framework comprises three main components: (i) Cloth Part Classification Network (\textit{PC-Net}) that identifies the grasped cloth region (edge, corner, inner fabric, or grasp failure), (ii) SAM-backboned Mask Decoder Network (\textit{SD-Net}) that synthesizes realistic tactile images to alleviate data scarcity, and (iii) Edge Pose Estimation Network (\textit{PE-Net}) that estimates edge center position and orientation for sliding control. In this work, we formulate the manipulation policy as a camera-free visuotactile control problem. All state feedback used for decision making is derived from the on-finger DIGIT sensors, with no external cameras used during operation. This design choice reflects the frequent self-occlusions encountered during edge tracing, where global vision becomes unreliable.

\subsection{Cloth Part Classification Network}
The Cloth Part Classification Network ({\it PC-Net}) assigns each visuotactile observation to one of four classes:

\begin{itemize}[leftmargin=10pt, label=•]
    \item \textit{Edge:} the sensor is in contact with a cloth edge.
    \item \textit{Corner:} the sensor is in contact with a cloth corner.
    \item \textit{In-Fabric:} the contact lies in the interior region.
    \item \textit{Grasp Failure:} no cloth is present in the sensing area.
\end{itemize}

This information is used to decide when to initiate sliding, when to re-grasp, and when a sliding trajectory has successfully reached the opposite corner.

\subsubsection{Data Preparation and Augmentation}
To obtain a balanced dataset without excessive data collection, we perform 100 grasp trials per class on multiple fabrics. Each trial produces a short sequence of tactile images. Standard 2D augmentation such as in-plane translation and rotation are applied, while depth augmentations are avoided to preserve temporal consistency between frames. After augmentation, the dataset contains on the order of 10,000 samples per class, covering a range of contact conditions, fabric patterns, and pose variations.

\subsubsection{Temporal Information for Classification}
Cloth contact evolves over time during a grasping motion. To exploit this temporal structure, \textit{PC-Net} processes a short sequence of frames rather than a single image. For each grasp attempt, the last five frames are stacked along a temporal dimension and passed through the network. By analyzing sequences of tactile images, the network can distinguish between transient and persistent features, improving its ability to classify complex interactions and adapt to varying fabric behaviors.

\subsubsection{Network Architecture}
\textit{PC-Net} adopts the Segment Anything Model (SAM) vision transformer backbone as a feature extractor, followed by a convolutional and fully connected head for the four-class classification task. Let $I \in \mathbb{R}^{B \times T \times C \times H \times W}$ denote a batch of $T$-frame tactile sequences. The SAM backbone performs:

\begin{itemize}[leftmargin=10pt]
    \item \textbf{Patch embedding:} each frame is partitioned into non-overlapping patches and projected into a $D$-dimensional feature space, yielding a set of patch tokens.
    \item \textbf{Transformer blocks:} a stack of self-attention blocks refines the patch tokens, capturing both local and global spatial dependencies.
    \item \textbf{Neck:} the refined tokens are reshaped into feature maps $I_{\text{neck}}$ suitable for convolutional processing.
\end{itemize}

The convolutional head then reduces the feature dimensionality while preserving spatial structure, using several convolutional layers with ReLU activation, batch normalization, and dropout. Temporal aggregation is implemented by averaging features across the $T$ frames, followed by a final fully connected layer that outputs logits in $\mathbb{R}^{B \times k}$ corresponding to the $k$ classes. This combination of transformer-based global context and convolutional local processing yields robust classification performance across fabrics and contact conditions.

\subsection{Edge Pose Estimation}
While \textit{PC-Net} provides a categorical description of the grasped region, successful sliding further requires precise estimation of edge pose within the tactile image. This subsection introduces the \textit{SD-Net} for synthetic data generation and the \textit{PE-Net} for edge pose estimation.

\subsubsection{SAM-backboned Mask Decoder Network}
Accurate edge pose estimation demands a diverse annotated tactile dataset, which is expensive to obtain manaully. To mitigate this, we design the \textit{SD-Net} to generate realistic tactile images from simple edge annotations.

\begin{figure}[t!]
    \centering
    \includegraphics[width=0.98\columnwidth]{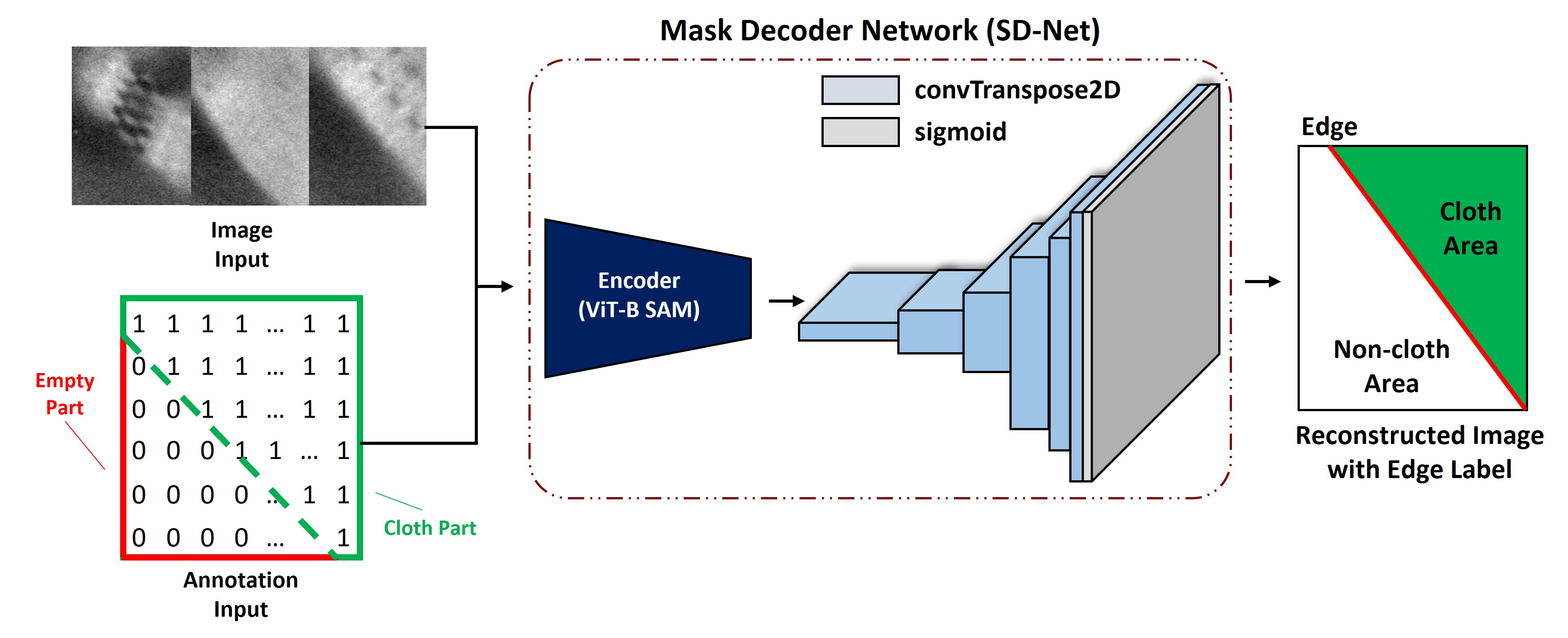}
    \caption{The architecture of SAM-backboned mask decoder network (\textit{SD-Net}). Network input is annotation information of real-world tactile images and output is generated tactile images compared with tactile image inputs.}
    \label{fig:Figure/MDN_figure.png}
\end{figure}

The architecture of \textit{SD-Net}, illustrated in Fig.~\ref{fig:Figure/MDN_figure.png}, integrates a SAM backbone with a convolutional decoder network to reconstruct tactile images from input masks. The \textit{SD-Net} includes:

\begin{itemize}[leftmargin=10pt, label=•]
\item \textbf{SAM backbone encoder:} the input mask (encoding the cloth region) is processed into a latent representation via patch embedding and transformer blocks, leveraging the strong spatial-context modeling of SAM.
\item \textbf{Convolutional decoder:} a stack of transposed convolutional layers progressively upsamples the latent representation to reconstruct a high-resolution tactile image.
\end{itemize}

Initially trained on a modest dataset comprising 100 manually annotated tactile edge images, \textit{SD-Net} is subsequently leveraged to generate 30,000 diverse synthetic images. This extensive augmentation significantly enrich the training distribution for pose estimation.

To enforce both pixel-wise accuracy and structural similarity, the training objective combines mean squared error (MSE) and structural similarity index measure (SSIM). The MSE loss between generated output $\hat{I}_0$ and ground-truth image $I_0$ is:

{\footnotesize
\begin{equation}
\mathcal{L}_{MSE} = \frac{1}{N} \sum_{i=1}^N \left( \hat{I}_{0,i} - I_{0,i} \right)^2
\end{equation}
}

\noindent
where $N$ is the number of pixels. For two image patches $x$ and $y$, SSIM is defined as:

{\footnotesize
\begin{equation}
SSIM(x,y) = \frac{(2\mu_x \mu_y + C_1)(2\sigma_{xy} + C_2)}{(\mu_x^2 + \mu_y^2 + C_1)(\sigma_x^2 + \sigma_y^2 + C_2)}
\end{equation}
}

where \(\mu_x\) and \(\mu_y\) are the mean intensities, \(\sigma_x^2\) and \(\sigma_y^2\) are variances, \(\sigma_{xy}\) is the covariance, and \(C_1\), \(C_2\) are constants for numerical stability. The SSIM loss is then:

{\footnotesize
\begin{equation}
\mathcal{L}_{SSIM} = 1 - SSIM(x,y)
\end{equation}
}

The total loss is a weighted sum of these components:

{\footnotesize
\begin{equation}
\mathcal{L} = \alpha \cdot \mathcal{L}_{MSE} + (1-\alpha) \cdot (1-\mathcal{L}_{SSIM})
\end{equation}
}

\noindent
where \(\alpha\) balances pixel-level accuracy and structural similarity. This training strategy enables \textit{SD-Net} to generate structurally and visually consistent tactile images from minimal inputs, providing a scalable way to enrich the pose-estimation dataset.

\subsubsection{Cloth Edge Pose Estimation Network}

Building on the outputs of \textit{SD-Net}, we train the \textit{PE-Net} to estimate the center position (\(x\), \(y\)) and orientation (\(\theta\)) of fabric edges. The \textit{PE-Net} also uses the SAM backbone as a feature extractor, as in \textit{PC-Net}, but replaces the classification head with a regression head. he tactile image is passed through the backbone to obtain a compact feature map, which is then processed by a convolutional head and a fully connected layer. The final output has dimension $\mathbb{R}^{B \times 3}$, corresponding to $(x,y,\theta)$ for each sample in the batch.

Training data is collected from sliding interactions between the gripper and multiple fabrics of various patterns and sizes. Approximately 200 real tactile images with labeled edge poses are obtained and then combined with the synthetic samples generated by \textit{SD-Net}. This hybrid dataset increases coverage in edge orientation, position, and fabric appearance, reducing overfitting and improving generalization.

The \textit{PE-Net} is trained using a combination of two loss functions: MSE loss for position prediction and angular loss for orientation prediction. The MSE loss ensures accurate localization of the edge center, while the angular loss minimizes deviations in the predicted orientation. For the angular component, we employ a cosine similarity loss that treats angles as vectors on a unit circle:

{\footnotesize
\begin{equation}
\mathcal{L}_{\text{angular}} = 1 - \cos\left(\theta_{\text{pred}} - \theta_{\text{true}}\right)
\end{equation}
}

This loss formulation is particularly effective in handling the periodic nature of angular measurements and ensures that small angular discrepancies are appropriately penalized. The total loss function is:

{\footnotesize
\begin{equation}
\mathcal{L} = \lambda_1 \cdot \mathcal{L}_{MSE} + \lambda_2 \cdot \mathcal{L}_{angular}
\end{equation}
}

\noindent
where \(\lambda_1\) and \(\lambda_2\) are weighting factors to balance positional and angular accuracy.

\begin{figure}[t]
    \centering
    \includegraphics[width=0.98\columnwidth]{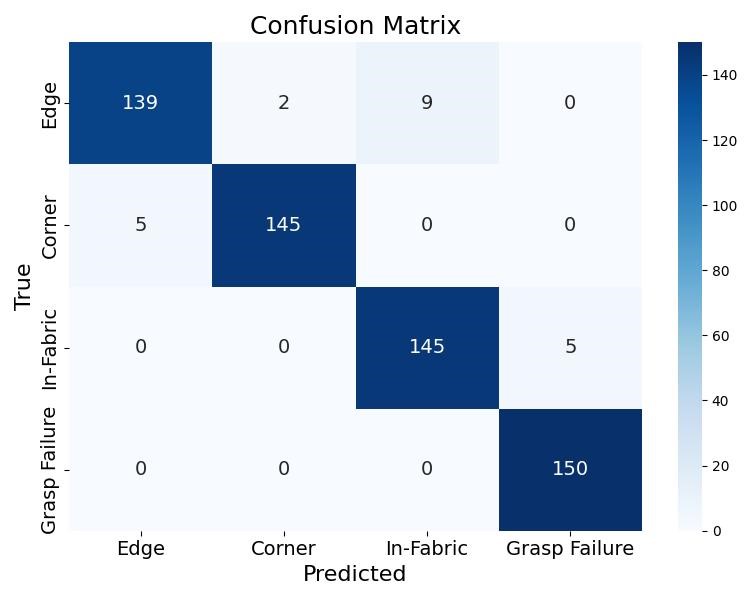}
    \caption[Confusion matrix of predicted vs. true labels for four classes—(1) Edge, (2) Corner, (3) In‐Fabric, and (4) Grasp Failure.]{Confusion matrix of predicted vs. true labels for four classes—(1) Edge, (2) Corner, (3) In‐fabric, and (4) Grasp Failure. Each cell shows the number of instances classified into each category, with diagonal cells indicating correct predictions.}
    \label{fig:CPCN_matrix}
\end{figure}

\begin{table*}[t]
\centering
\caption{Classification evaluation for \textit{PC-Net}.}
\small
\renewcommand{\arraystretch}{1.3}

\begin{tabular}{c | C{1.5cm} C{1.5cm} C{1.5cm} C{1.5cm} | C{1.5cm} C{1.5cm} C{1.5cm} C{1.5cm}}
\specialrule{1.3pt}{0.0pt}{1pt}
\multirow{2}{*}[-0.5em]{\textbf{Method / Metric}} & \multicolumn{4}{c|}{\bf Precision (\%)} & \multicolumn{4}{c}{\bf Recall (\%)}\\ [0.2ex]
\cline{2-9}
& \makecell[c]{\rule{0pt}{3ex}Edge}
& \makecell[c]{\rule{0pt}{3ex}Corner} 
& \makecell[c]{\rule{0pt}{3ex}In-Fabric} 
& \makecell[c]{Grasp\\Failure}
& \makecell[c]{\rule{0pt}{3ex}Edge} 
& \makecell[c]{\rule{0pt}{3ex}Corner} 
& \makecell[c]{\rule{0pt}{3ex}In-Fabric} 
& \makecell[c]{Grasp\\Failure} \\
\hline
B1: ResNet     &0.85 &0.89 &0.94 &0.96 &0.85 &0.84 &0.95 &1.00 \\
\hline
B2: DenseNet   &0.92 &0.92 &{\textbf{0.95}} &0.97 &0.91 &0.89 &0.97 &1.00 \\
\hline
B3: ViT        &0.87 &0.81 &0.79 &0.90 &0.69 &0.83 &0.87 &0.98 \\
\hline
\rowcolor{gray!20}
{\textbf{PC-Net (Ours)}} &{\textbf{0.97}} &{\textbf{0.99}} &0.94 &{\textbf{0.97}} &{\textbf{0.93}} &{\textbf{0.97}} &{\textbf{0.97}} &{\textbf{1.00}} \\
\specialrule{1.3pt}{0.0pt}{1pt}
\end{tabular}

\label{table:CPCN_classification}
\end{table*}

\begin{figure}[b]
    \centering
    \includegraphics[width=0.98\columnwidth]{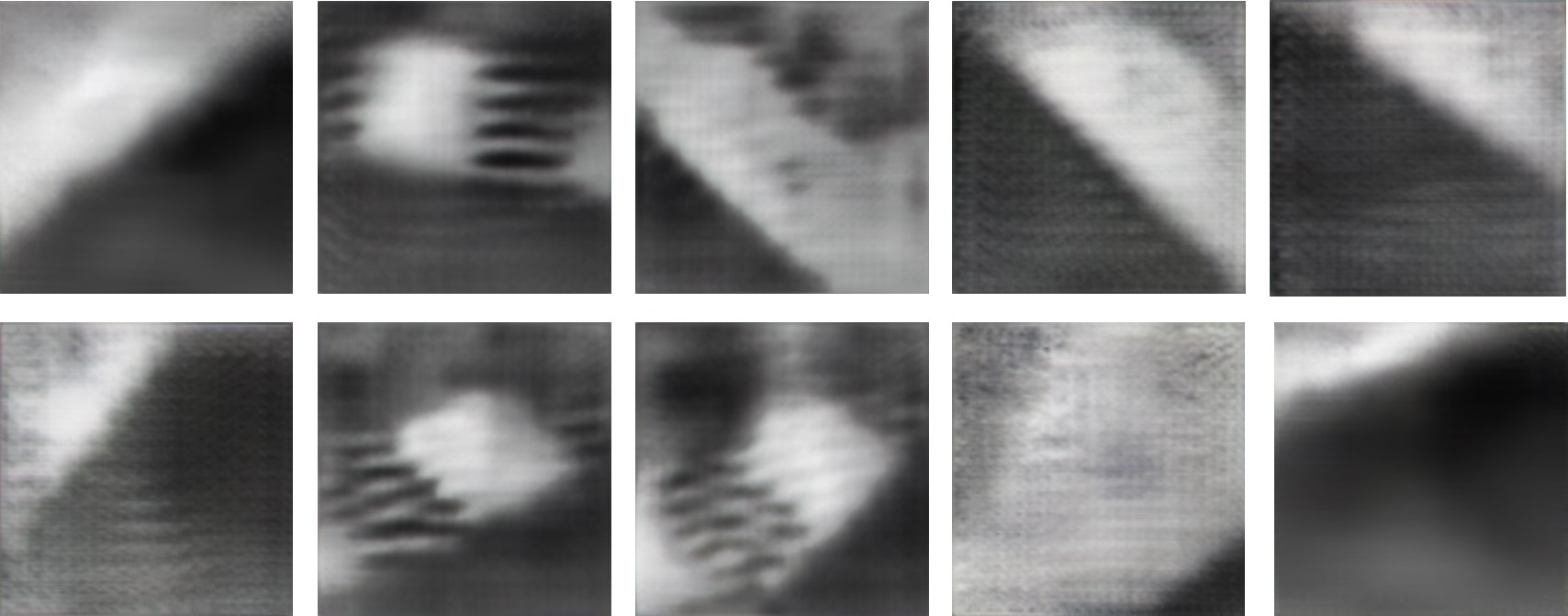}
    \caption[The sample outputs of SAM-backboned mask decoder network (\textit{SD-Net}).]{The sample outputs of SAM-backboned mask decoder network (\textit{SD-Net}). When generating images, random edge pose is input to create synthetic images for random edge poses. This enables to get unseen images that is not captured when training.}
    \label{fig:sam_mdn_imgs}
\end{figure}

By combining the synthetic data generated by \textit{SD-Net} with a small set of real-world data, we achieve precise edge pose predictions. These predictions are subsequently used to guide real-time tactile sliding for tasks such as edge tracing. 

\begin{figure}[t]
    \centering
    \includegraphics[width=0.98\columnwidth]{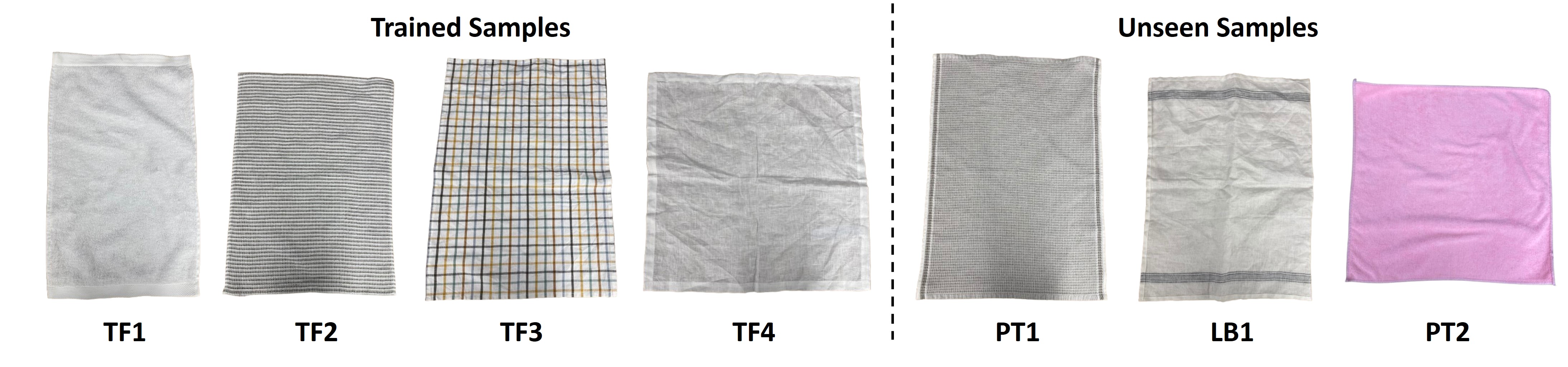}
    \caption{Cloth samples used to evaluate the generalization capability of the proposed cloth manipulation framework.}
    \label{fig:tested_cloths}
\end{figure}

\section{Experiments} \label{sec:experiments}
We evaluate the proposed perception and manipulation framework in four steps: (i) classification performance of \textit{PC-Net} compared with standard deep-learning baselines, (ii) analysis of the synthetic tactile data generated by \textit{SD-Net}, (iii) quantitative edge pose estimation accuracy of \textit{PE-Net}, and (iv) real-world cloth sliding performance using the \textit{Touch G.O.G.} system.

\subsection{\textit{PC-Net} Evaluation}
The \textit{PC-Net} model is evaluated on four trained cloths and three unseen cloths, as illustrated in Fig.~\ref{fig:tested_cloths}. We compare its cloth-part classification performance against ResNet50, DenseNet121, and Vision Transformer (ViT) backbones. Table~\ref{table:CPCN_classification} reports precision and recall for the four classes (\textit{Edge}, \textit{Corner}, \textit{In-Fabric}, \textit{Grasp Failure}).

All methods achieve similar performance on the \textit{Grasp Failure}, but \textit{PC-Net} clearly improves the recognition of \textit{Edge} and \textit{Corner}. This is particularly important for the sliding pipeline. Accurate detection of edge is used to trigger the \textit{PE-Net} and reliable corner recognition is required at the start and end of a sliding motion to ensure correct cloth alignment.

The confusion matrix in Fig.~\ref{fig:CPCN_matrix} further illustrates the performance. Most samples lie on the main diagonal for all four classes, indicating consistent behavior across different fabrics. The \textit{Edge} class exhibits minor confusion with \textit{In-Fabric} (9 samples) and \textit{Corner} (2 samples), but the overall classification accuracy remains high. In practice, these results indicate that standard data augmentation on real tactile sequences is sufficient for this task, meaning additional synthetic data is not required for \textit{PC-Net}.

\begin{figure}[t]
    \centering
    \includegraphics[width=0.98\columnwidth]{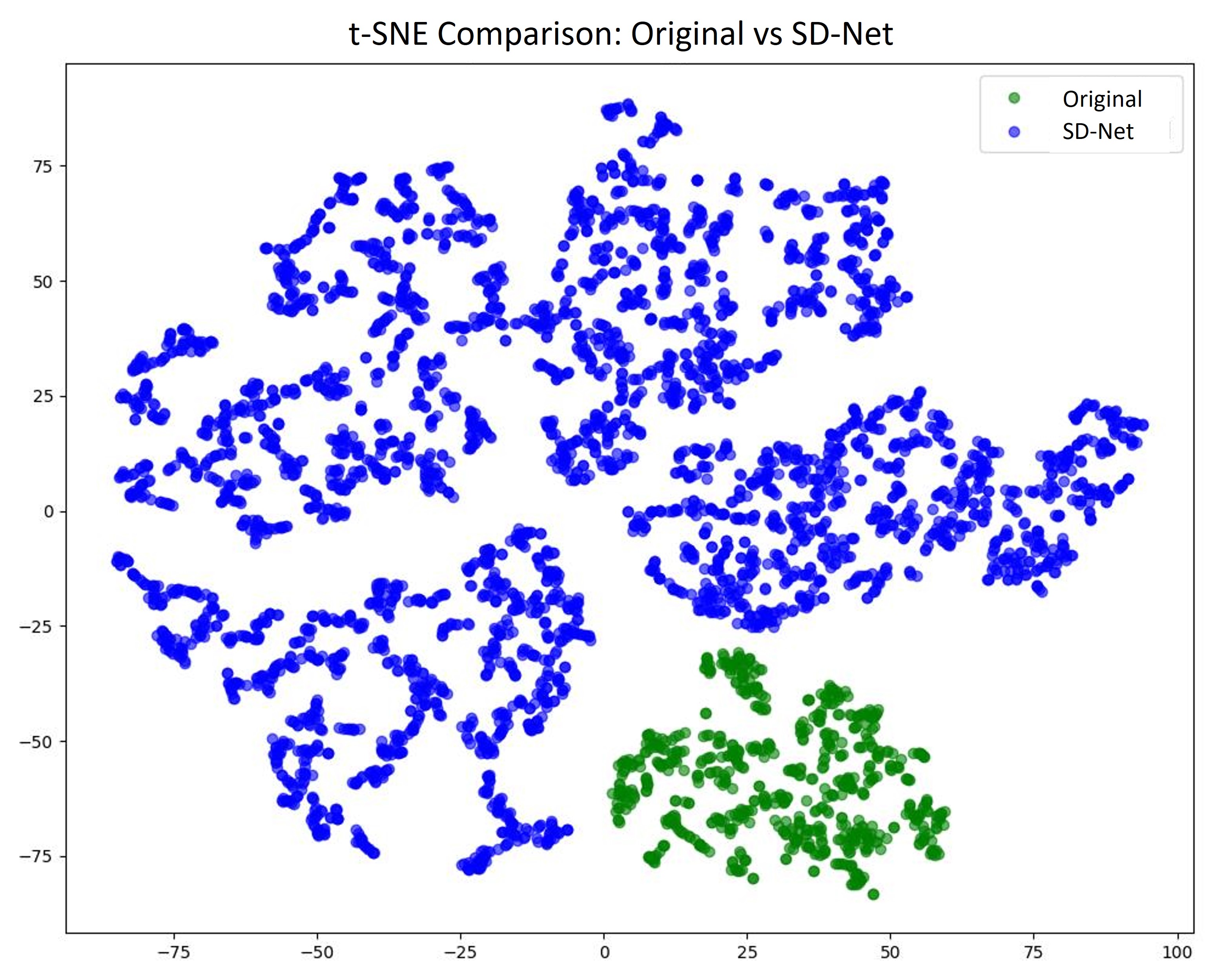}
    \caption[t-SNE visualization comparing \textit{SD-Net}-generated data with the original real-world dataset.]{t-SNE visualization comparing \textit{SD-Net}-generated data (blue) with the real tactile dataset (green). The synthetic data exhibits broader coverage, indicating improved generalization and robustness.}
    \label{fig:t_sne_analysis}
\end{figure}

\begin{table}[b]
\centering
\caption{Edge pose estimation assessment for \textit{PE-Net}.}
\renewcommand{\arraystretch}{1.2}
\setlength{\tabcolsep}{3mm}
\begin{tabular}{c | c c c | c}
\specialrule{1.3pt}{0.0pt}{1pt}
\multirow{2}{*}{\textbf{Method / Metric}} & \multicolumn{3}{c|}{\bf Position (mm)} & \multirow{2}{*}{\makecell{\bf Angle \bf (°)}} \\ [0.2ex]
\cline{2-4}
& {\bf X} & {\bf Y} & {\bf Distance} & \\
\hline
Edge Detection     & 2.47 & 2.32 & 3.38 & 50.22 \\
\hline
ResNet-w/o &0.68 &1.58 &1.72 &14.92 \\
\hline
DenseNet-w/o &0.46 &1.13 &1.22 &8.19 \\
\hline
ViT-w/o &0.37 &0.85 &0.93 &6.48 \\
\hline
PE-Net-w/o &0.30 &0.72 &0.78 &5.52 \\
\specialrule{1.3pt}{0.0pt}{0.0pt} 
ResNet     & 0.44 & 0.88 & 0.99 & 12.55 \\
\hline
DenseNet  & 0.41 & 0.96 & 1.04 & 7.03 \\
\hline
ViT       & 0.28 & 0.78 & 0.83 & 5.35 \\
\hline
\rowcolor{gray!20}
\textbf{PE-Net (Ours)}    & {\bf 0.25} & {\bf 0.54} & {\bf 0.59} & {\bf 4.52} \\
\specialrule{1.3pt}{0.0pt}{1pt}
\end{tabular}
\label{table:edge_pose_assessment}
\end{table}

\begin{figure*}[t]
    \centering
    \includegraphics[width=0.98\linewidth]{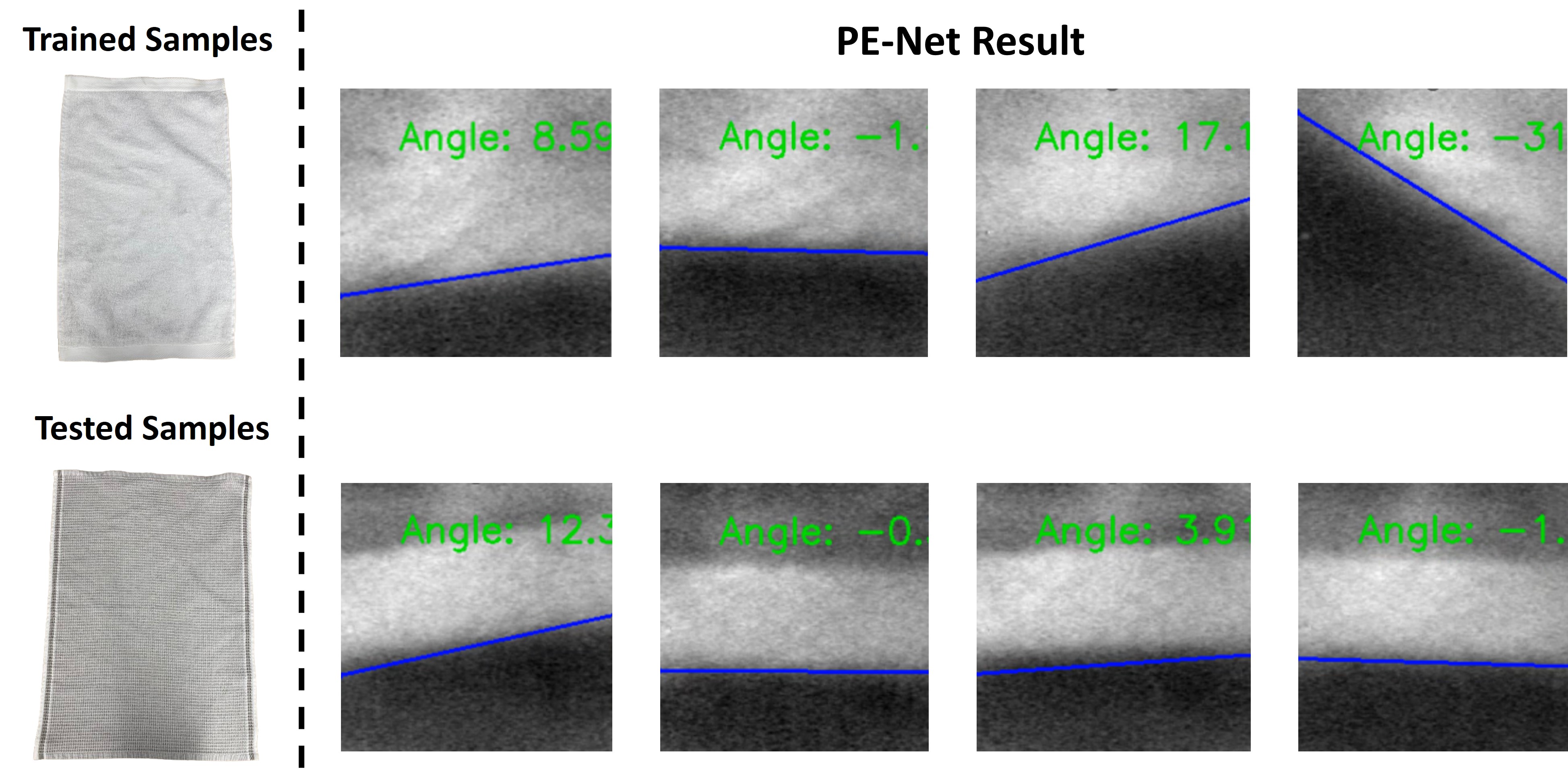}
    \caption[Example results of the \textit{PE-Net} algorithm estimating the edge center position and angle for both trained and tested fabric samples.]{Example results of the \textit{PE-Net} edge pose estimation on trained (top row) and unseen (bottom row) fabric samples. Each sub-image shows the estimated edge (blue line) and the predicted angle (green text), illustrating consistent boundary detection and orientation estimation under diverse conditions.}
    \label{fig:epen_sample_result}
\end{figure*}

\subsection{\textit{SD-Net} Evaluation}
In contrast to classification, edge pose estimation requires substantially more labeled variability in edge positions and orientations than can be obtained from real data alone. Here, \textit{SD-Net} is used to generate a large synthetic dataset that complements the limited set of manually annotated tactile images. Figure~\ref{fig:sam_mdn_imgs} shows representative examples of images synthesized by \textit{SD-Net} from randomly sampled edge annotations. The generated images capture key characteristics of real visuotactile observations, including contact footprint shape and local texture, while enabling systematic variation in edge pose beyond what is accessible in the data collection stage.

To evaluate the effectiveness of the \textit{SD-Net} in capturing the underlying data distribution, we perform a t-SNE analysis on latent features extracted from both the real and synthetic datasets. As shown in Fig.~\ref{fig:t_sne_analysis}, the \textit{SD-Net}-generated samples (blue) occupy a broader region in the embedded space compared to the real images (green), which are confined to a smaller subset. This indicates that \textit{SD-Net} learns a richer representation of plausible edge configurations and effectively interpolates between real examples.

From the control perspective, this broader coverage is beneficial. The pose estimator is exposed during training to a wider range of edge orientations and contact patterns, including configurations that are rare in practice but can occur during sliding. Moreover, the smooth distribution of synthetic samples reduces discontinuities in the feature space, which is advantageous for downstream learning-based estimation in the presence of sensor noise and fabric variability. Overall, the analysis confirms that \textit{SD-Net} provides a useful and physically consistent augmentation of the real tactile dataset, forming a crucial component of the edge pose estimation pipeline described in the following sections.

\begin{figure*}[t]
    \centering
    \includegraphics[width=\linewidth]{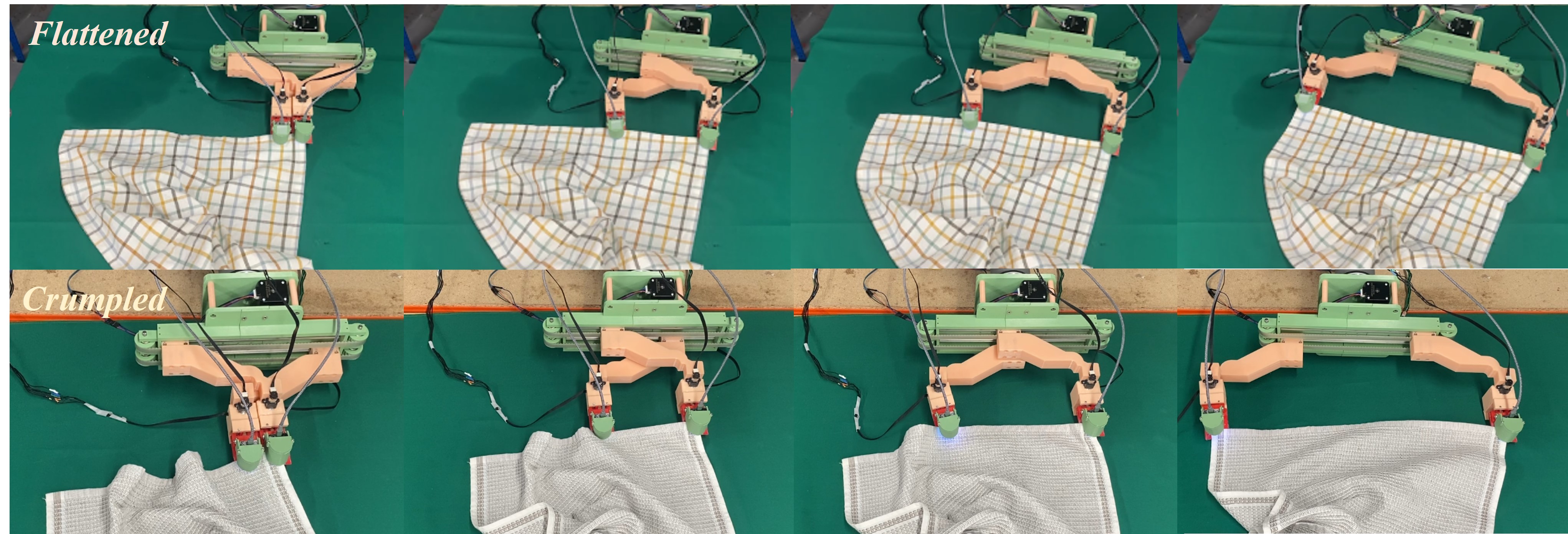}
    \caption{Real-world cloth sliding experiment using the {\it Touch G.O.G.} system. The rows illustrate sequential sliding steps for the flattened and crumpled fabrics. One T-VFG initially grasps a cloth corner, while the other performs sliding until the opposite corner is detected. During sliding, the gripper adjusts its opening width and abduction angle based on edge pose estimates provided by \textit{PE-Net}.}
    \label{fig:tgog_slide}
\end{figure*}

\subsection{\textit{PE-Net} Evaluation}
Table~\ref{table:edge_pose_assessment} reports quantitative edge pose estimation performance for several baselines (classical edge detection~\cite{harris1988combined}, ResNet, DenseNet, ViT) and the proposed \textit{PE-Net}, with and without synthetic data from \textit{SD-Net} (denoted “w/o”). Position errors are reported along $x$, $y$, and as Euclidean distance in millimeters, together with angular error in degrees.

As shown in Table~\ref{table:edge_pose_assessment}, \textit{PE-Net} achieves the lowest position and angle errors among all methods, with an average distance error of $0.59\,\mathrm{mm}$ and angular error of $4.52^\circ$. This represents a substantial improvement over classical edge detection ($3.38\,\mathrm{mm}$, $50.22^\circ$) and also over the best deep-learning baseline (ViT, $0.83\,\mathrm{mm}$, $5.35^\circ$). Such sub-millimeter localization and low orientation error are critical for maintaining the cloth edge centered on the tactile sensor and aligning the gripper opening with the edge during sliding.

The comparison between \textit{PE-Net} and \textit{PE-Net-w/o} highlights the role of synthetic data generated by \textit{SD-Net}. Training without synthetic data leads to a higher distance error of $0.78\,\mathrm{mm}$ and angular error of $5.52^\circ$, whereas incorporating \textit{SD-Net} samples reduces these to $0.59\,\mathrm{mm}$ and $4.52^\circ$, respectively. This confirms that the broader pose coverage provided by \textit{SD-Net} (see Fig.~\ref{fig:sam_mdn_imgs} and the t-SNE analysis in Fig.~\ref{fig:t_sne_analysis}) directly translates into improved estimation accuracy.

Figure~\ref{fig:epen_sample_result} illustrates qualitative results for both trained and unseen fabrics. In all cases, the estimated edge (blue line) closely tracks the visible boundary in the tactile image, and the predicted orientation remains stable despite changes in texture, pattern, and local deformation. This robustness to fabric variability is essential for deployment in the \textit{Touch G.O.G.} sliding controller, where estimated edge poses are used to adjust the abduction angle and yaw of the gripper. Overall, the experiments demonstrate that \textit{PE-Net}, combined with \textit{SD-Net} augmentation, provides the accuracy and generalization required for reliable single-arm bimanual cloth manipulation.

\subsection{Real-world Cloth Sliding Evaluation}
We evaluate the proposed visuotactile framework on real cloth sliding tasks using the \textit{Touch G.O.G.} mounted on a UR5 robotic arm, as illustrated in Fig.~\ref{fig:tgog_slide}. The objective is to assess how \textit{PC-Net} and \textit{PE-Net} jointly support single-arm bimanual unfolding. All control decisions rely exclusively on DIGIT tactile images with no reliance on external cameras, highlighting the strength of our tactile perception system.
 
The DIGIT sensors on the gripper capture tactile images, which are then processed by \textit{PC-Net} to identify contact regions (e.g., corner and edge) and by \textit{PE-Net} to estimate the edge center and orientation when an edge is detected. Once the grip region and edge pose are determined, the \textit{Touch G.O.G.} attempts a sliding motion on the fabric. Due to variability in fabric properties and configurations, some fabric orientations may exceed the inherent sliding capabilities of the \textit{Touch G.O.G.}, requiring adjustments by the robotic arm. 

\begin{figure}[t]
    \centering
    \includegraphics[width=0.98\columnwidth]{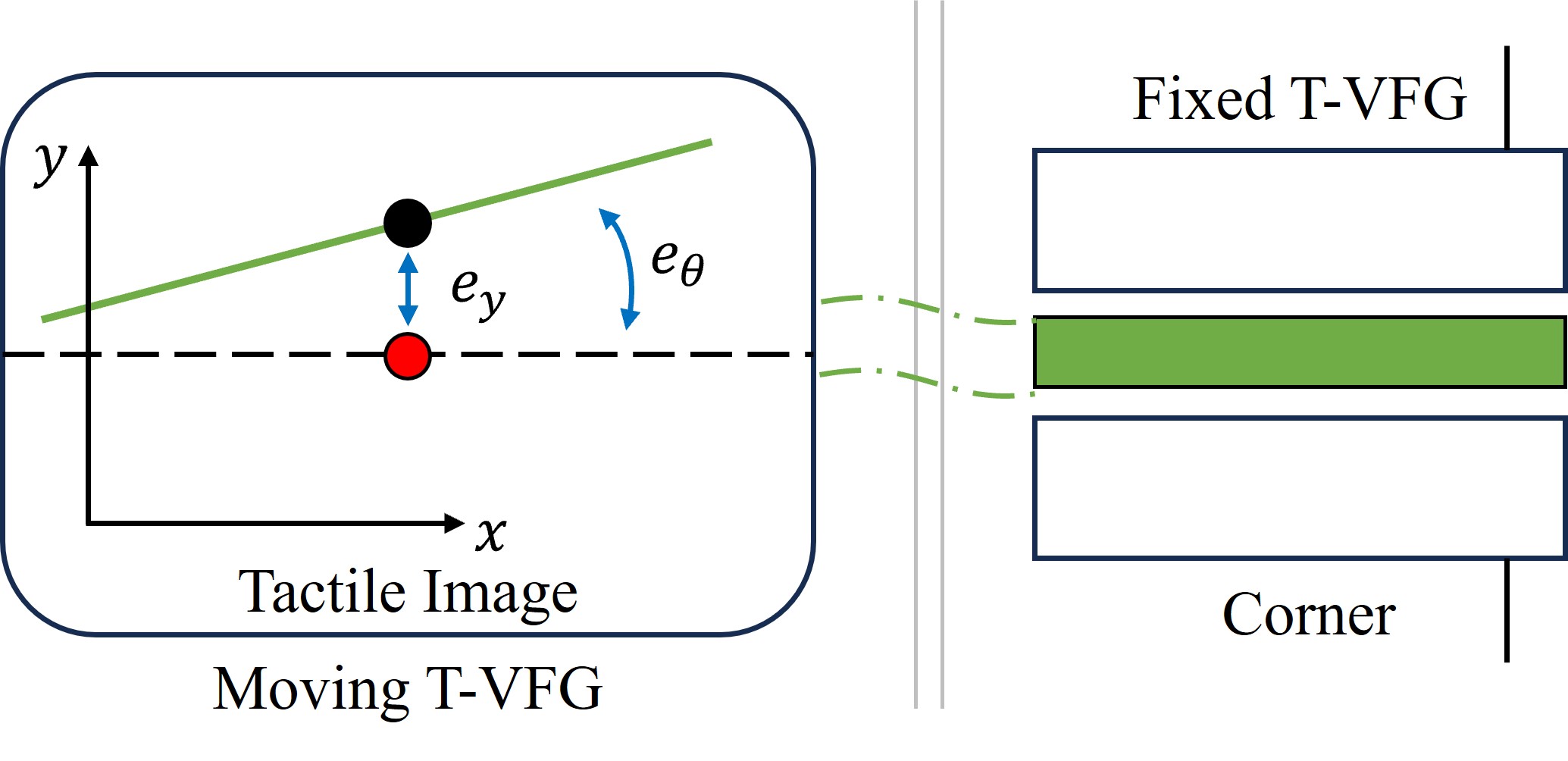}
    \caption{Visuotactile edge-alignment control for cloth sliding. The fixed T-VFG holds a cloth corner, while the moving T-VFG uses the DIGIT image to regulate $e_y$ and $e_{\theta}$ during sliding.}
    \label{fig:control_scheme}
\end{figure}

When the T-VFGs slide along a fabric edge, the detected edge needs to stay centered on the DIGIT sensing area and aligned with the $x$-axis of the image, as illustrated in Fig.~\ref{fig:control_scheme}. Let $e_{y}(k)$ denote the vertical offset of the detected edge from the center of the image at time step $k$, and $e_{\theta}(k)$ the angular misalignment between the estimated edge orientation and the $x$-axis. Two discrete PD controllers are used:

{\footnotesize
\begin{equation}
  u_{yaw}(k) \;=\; K_{P_y}\,e_y(k) 
          \;+\; K_{D_y}\,\frac{e_y(k) - e_y(k-1)}{\Delta T}
\label{eq:pd_rotation}
\end{equation}

\begin{equation}
  u_{ab}(k) \;=\; K_{P_{\theta}}\,[e_{\theta}(k)-\beta u_{yaw}(k)] 
          \;+\; K_{D_{\theta}}\,\frac{e_{\theta}(k) - e_{\theta}(k-1)}{\Delta T}
\label{eq:pd_abduction}
\end{equation}
}

\noindent
where $u_{\text{yaw}}$ commands the end-effector yaw of the UR5, $u_{\text{ab}}$ commands the abduction angle of the T-VFG, and $\Delta T$ is the control interval. The feed-forward term $\beta u_{\text{yaw}}$ compensates for the coupling between end-effector rotation and the perceived edge tilt, and both commands are clipped to $u_{\text{yaw}}\in[-5^\circ,5^\circ]$ and $u_{\text{ab}}\in[-30^\circ,30^\circ]$. This scheme keeps the estimated edge centered in the DIGIT image and aligned with the sensor $x$-axis.

\begin{table}[b]
\centering
\caption{Performance of Cloth Sliding}

\setlength{\tabcolsep}{1.7mm}{
\begin{tabular}{c|c c c c c c c| c}
\specialrule{1.3pt}{0.5pt}{1pt}
\textbf{Config.}  & \textbf{TF1} & \textbf{TF2} & \textbf{TF3} & \textbf{TF4} & \textbf{PT1} & \textbf{LB1} & \textbf{PT2} & \textbf{Average} \\ [0.2ex]
\hline

\makecell[c]{\emph{Flattened}} & 4/5 & 3/5 & 3/5 & 4/5 & 3/5 & 4/5 & 3/5 & 24/35 \\
\hline

 \makecell[c]{\emph{Crumpled}} & 3/5 & 2/5 & 2/5 & 4/5 & 3/5 & 3/5 & 3/5 & 20/35 \\
\specialrule{1.3pt}{0.5pt}{1pt}

\end{tabular}
}

\par
\begin{flushleft}
Seven different fabrics were tested under two initial cloth configurations: \emph{flattened} and \emph{crumpled}. The scores represent the number of successful trials out of five attempts, illustrating the robustness of our framework across diverse materials and initial conditions.

\end{flushleft}
\label{table:sliding_success}
\end{table}

In addition to edge pose control, discrete adjustments of the insertion depth are triggered by \textit{PC-Net}. When the sliding T-VFG is classified as \textit{In-Fabric}, indicating that the gripper is too deeply embedded to slide effectively, the UR5 rotates slightly clockwise and withdraws a small distance, resulting in a shallower insertion. Conversely, when the sliding T-VFG is classified as \textit{Grasp Failure}, the UR5 rotates slightly counter-clockwise and inserts the gripper deeper, increasing fabric engagement. Sliding continues with these small corrections until both T-VFGs are classified as \textit{Corner}, signaling that the opposite corner has been reached.

To isolate the contribution of tactile perception, we constrain the setup so that one T-VFG always starts from a correctly grasped corner and no global vision is used for corner detection or pose correction. This allows us to directly evaluate how well \textit{PC-Net} and \textit{PE-Net} support local edge tracking and unfolding based solely on visuotactile feedback.

As illustrated in Fig.~\ref{fig:tested_cloths}, we test seven fabrics with different patterns, thicknesses, and material properties, including four template fabrics (TF1–TF4), a patterned towel (PT1), a linen blanket (LB1), and a pink towel (PT2). Each fabric is evaluated under two initial configurations: \emph{flattened} (roughly planar configuration) and \emph{crumpled} (randomly compressed configuration). For each combination, we perform five trials, resulting in a total of 70 experiments. A trial is counted as successful if the sliding T-VFG reaches the opposite corner and both T-VFGs are classified as \textit{Corner} at the end of the motion, with the cloth sufficiently unfolded for subsequent handling.

Table~\ref{table:sliding_success} summarizes the sliding success rates. The flattened configuration, which presents simpler edge geometries, achieves 24/35 successful trials. The crumpled configuration is more challenging due to folds and non-uniform thickness, yet still yields 20/35 successful trials across all fabrics. These results indicate that the combination of \textit{PC-Net} and \textit{PE-Net}, together with the \textit{Touch G.O.G.} system, can handle substantial variability in fabric pattern and initial configuration while relying solely on local visuotactile feedback.

Overall, the experiments confirm that the proposed framework enables robust edge tracing with a single robotic arm, even for patterned fabrics and crumpled initial states. By relying solely on visuotactile feedback, the system demonstrates robust performance even in scenarios where external vision may be compromised. These findings underscore the potential of our approach for reliable, uncertainty-aware cloth manipulation in practical robotic applications.

\section{Conclusion} \label{sec:conclusion}

This paper presents \textit{Touch G.O.G.}, a single-arm robot system that achieves bimanual cloth manipulation by combining a novel visuotactile gripper with dedicated perception and control algorithms. The end-effector comprises a Decoupled Width Control Gripper (D-WCG) with independently actuated, belt-driven fingers and Tactile Variable Friction Grippers (T-VFGs) equipped with vision-based tactile sensors and DC motors. This design enables large in-plane span, abduction-based in-gripper reorientation, and regulated contact forces suitable for sliding along cloth edges under occlusion.

On the perception side, we introduce a SAM-based cloth part classification network (\textit{PC-Net}) and edge pose estimation network (\textit{PE-Net}), together with a SAM-backboned mask decoder network (\textit{SD-Net}) that generates synthetic tactile images to alleviate data scarcity. The resulting visuotactile pipeline reliably classifies edges, corners, inner-fabric regions, and grasp failures, and achieves sub-millimeter edge localization with low angular error. These estimates are used to maintain the edge centered and aligned in the tactile image during sliding.

Real-world experiments on multiple fabrics, including both flattened and crumpled initial configurations, demonstrate that the proposed system can robustly slide from corner to corner using only local tactile feedback, without external vision. The results highlight the effectiveness of combining a visuotactile gripper with foundation-model–based perception and synthetic data generation for deformable object manipulation.

Future work will focus on scaling to larger garments, integrating additional sensing modalities (e.g., force–torque or global vision) for multi-step tasks such as folding or dressing, and assessing with different robotic platforms. We anticipate that these developments will further expand the applicability of visuotactile grippers such as \textit{Touch G.O.G.} in domestic, industrial, and healthcare environments.



\bibliographystyle{IEEEtran}
\bibliography{references}

@inproceedings{harris1988combined,
  title={A combined corner and edge detector},
  author={Harris, Chris and Stephens, Mike and others},
  booktitle={Alvey vision conference},
  volume={15},
  number={50},
  pages={10--5244},
  year={1988},
  organization={Manchester, UK}
}

@article{jimenez2020perception,
  title={Perception of cloth in assistive robotic manipulation tasks},
  author={Jim{\'e}nez, Pablo and Torras, Carme},
  journal={Natural Computing},
  volume={19},
  number={2},
  pages={409--431},
  year={2020},
  publisher={Springer}
}

@inproceedings{martinez2018recognition,
  title={Recognition of grasp points for clothes manipulation under unconstrained conditions},
  author={Mart{\'\i}nez, Luz Mar{\'\i}a and Ruiz-del-Solar, Javier},
  booktitle={RoboCup 2017: Robot World Cup XXI 11},
  pages={350--362},
  year={2018},
  organization={Springer}
}

@inproceedings{saxena2019garment,
  title={Garment recognition and grasping point detection for clothing assistance task using deep learning},
  author={Saxena, Krati and Shibata, Tomohiro},
  booktitle={2019 IEEE/SICE International Symposium on System Integration (SII)},
  pages={632--637},
  year={2019},
  organization={IEEE}
}

@inproceedings{seita2019deep,
  title={Deep transfer learning of pick points on fabric for robot bed-making},
  author={Seita, Daniel and Jamali, Nawid and Laskey, Michael and Tanwani, Ajay Kumar and Berenstein, Ron and Baskaran, Prakash and Iba, Soshi and Canny, John and Goldberg, Ken},
  booktitle={The International Symposium of Robotics Research},
  pages={275--290},
  year={2019},
  organization={Springer}
}

@inproceedings{jangir2020dynamic,
  title={Dynamic cloth manipulation with deep reinforcement learning},
  author={Jangir, Rishabh and Alenya, Guillem and Torras, Carme},
  booktitle={2020 IEEE International Conference on Robotics and Automation (ICRA)},
  pages={4630--4636},
  year={2020},
  organization={IEEE}
}

@article{chen2025graphgarment,
  title={GraphGarment: Learning Garment Dynamics for Bimanual Cloth Manipulation Tasks},
  author={Chen, Wei and Li, Kelin and Lee, Dongmyoung and Chen, Xiaoshuai and Zong, Rui and Kormushev, Petar},
  journal={arXiv preprint arXiv:2503.05817},
  year={2025}
}

@inproceedings{ha2022flingbot,
  title={Flingbot: The unreasonable effectiveness of dynamic manipulation for cloth unfolding},
  author={Ha, Huy and Song, Shuran},
  booktitle={Conference on Robot Learning},
  pages={24--33},
  year={2022},
  organization={PMLR}
}

@article{xu2022dextairity,
  title={Dextairity: Deformable manipulation can be a breeze},
  author={Xu, Zhenjia and Chi, Cheng and Burchfiel, Benjamin and Cousineau, Eric and Feng, Siyuan and Song, Shuran},
  journal={arXiv preprint arXiv:2203.01197},
  year={2022}
}

@article{yuba2017unfolding,
  title={Unfolding of a rectangular cloth from unarranged starting shapes by a dual-armed robot with a mechanism for managing recognition error and uncertainty},
  author={Yuba, Hiroyuki and Arnold, Solvi and Yamazaki, Kimitoshi},
  journal={Advanced Robotics},
  volume={31},
  number={10},
  pages={544--556},
  year={2017},
  publisher={Taylor \& Francis}
}

@article{gabas2021dual,
  title={Dual edge classifier for robust cloth unfolding},
  author={Gabas, Antonio and Kita, Yasuyo and Yoshida, Eiichi},
  journal={ROBOMECH Journal},
  volume={8},
  number={1},
  pages={15},
  year={2021},
  publisher={Springer}
}

@article{garcia2020benchmarking,
  title={Benchmarking bimanual cloth manipulation},
  author={Garcia-Camacho, Irene and Lippi, Martina and Welle, Michael C and Yin, Hang and Antonova, Rika and Varava, Anastasiia and Borras, Julia and Torras, Carme and Marino, Alessandro and Alenya, Guillem and others},
  journal={IEEE Robotics and Automation Letters},
  volume={5},
  number={2},
  pages={1111--1118},
  year={2020},
  publisher={IEEE}
}

@inproceedings{yuan2018active,
  title={Active clothing material perception using tactile sensing and deep learning},
  author={Yuan, Wenzhen and Mo, Yuchen and Wang, Shaoxiong and Adelson, Edward H},
  booktitle={2018 IEEE International Conference on Robotics and Automation (ICRA)},
  pages={4842--4849},
  year={2018},
  organization={IEEE}
}

@article{huang2022understanding,
  title={Understanding dynamic tactile sensing for liquid property estimation},
  author={Huang, Hung-Jui and Guo, Xiaofeng and Yuan, Wenzhen},
  journal={arXiv preprint arXiv:2205.08771},
  year={2022}
}

@inproceedings{tian2019manipulation,
  title={Manipulation by feel: Touch-based control with deep predictive models},
  author={Tian, Stephen and Ebert, Frederik and Jayaraman, Dinesh and Mudigonda, Mayur and Finn, Chelsea and Calandra, Roberto and Levine, Sergey},
  booktitle={2019 International Conference on Robotics and Automation (ICRA)},
  pages={818--824},
  year={2019},
  organization={IEEE}
}

@article{lambeta2020digit,
  title={Digit: A novel design for a low-cost compact high-resolution tactile sensor with application to in-hand manipulation},
  author={Lambeta, Mike and Chou, Po-Wei and Tian, Stephen and Yang, Brian and Maloon, Benjamin and Most, Victoria Rose and Stroud, Dave and Santos, Raymond and Byagowi, Ahmad and Kammerer, Gregg and others},
  journal={IEEE Robotics and Automation Letters},
  volume={5},
  number={3},
  pages={3838--3845},
  year={2020},
  publisher={IEEE}
}

@inproceedings{kolamuri2021improving,
  title={Improving grasp stability with rotation measurement from tactile sensing},
  author={Kolamuri, Raj and Si, Zilin and Zhang, Yufan and Agarwal, Arpit and Yuan, Wenzhen},
  booktitle={2021 IEEE/RSJ International Conference on Intelligent Robots and Systems (IROS)},
  pages={6809--6816},
  year={2021},
  organization={IEEE}
}

@inproceedings{taylor2022gelslim,
  title={Gelslim 3.0: High-resolution measurement of shape, force and slip in a compact tactile-sensing finger},
  author={Taylor, Ian H and Dong, Siyuan and Rodriguez, Alberto},
  booktitle={2022 International Conference on Robotics and Automation (ICRA)},
  pages={10781--10787},
  year={2022},
  organization={IEEE}
}

@article{lin2022tactile,
  title={Tactile gym 2.0: Sim-to-real deep reinforcement learning for comparing low-cost high-resolution robot touch},
  author={Lin, Yijiong and Lloyd, John and Church, Alex and Lepora, Nathan F},
  journal={IEEE Robotics and Automation Letters},
  volume={7},
  number={4},
  pages={10754--10761},
  year={2022},
  publisher={IEEE}
}

@inproceedings{tirumala2022learning,
  title={Learning to singulate layers of cloth using tactile feedback},
  author={Tirumala, Sashank and Weng, Thomas and Seita, Daniel and Kroemer, Oliver and Temel, Zeynep and Held, David},
  booktitle={2022 IEEE/RSJ International Conference on Intelligent Robots and Systems (IROS)},
  pages={7773--7780},
  year={2022},
  organization={IEEE}
}

@article{she2021cable,
  title={Cable manipulation with a tactile-reactive gripper},
  author={She, Yu and Wang, Shaoxiong and Dong, Siyuan and Sunil, Neha and Rodriguez, Alberto and Adelson, Edward},
  journal={The International Journal of Robotics Research},
  volume={40},
  number={12-14},
  pages={1385--1401},
  year={2021},
  publisher={SAGE Publications Sage UK: London, England}
}

@inproceedings{pecyna2022visual,
  title={Visual-tactile multimodality for following deformable linear objects using reinforcement learning},
  author={Pecyna, Leszek and Dong, Siyuan and Luo, Shan},
  booktitle={2022 IEEE/RSJ International Conference on Intelligent Robots and Systems (IROS)},
  pages={3987--3994},
  year={2022},
  organization={IEEE}
}

@inproceedings{sunil2023visuotactile,
  title={Visuotactile affordances for cloth manipulation with local control},
  author={Sunil, Neha and Wang, Shaoxiong and She, Yu and Adelson, Edward and Garcia, Alberto Rodriguez},
  booktitle={Conference on Robot Learning},
  pages={1596--1606},
  year={2023},
  organization={PMLR}
}

@inproceedings{zhou2021plas,
  title={Plas: Latent action space for offline reinforcement learning},
  author={Zhou, Wenxuan and Bajracharya, Sujay and Held, David},
  booktitle={Conference on Robot Learning},
  pages={1719--1735},
  year={2021},
  organization={PMLR}
}

@article{borras2022effective,
  title={Effective grasping enables successful robot-assisted dressing},
  author={Borr{\`a}s, J{\'u}lia},
  journal={Science robotics},
  volume={7},
  number={65},
  pages={eabo7229},
  year={2022},
  publisher={American Association for the Advancement of Science}
}

@article{zhang2023visual,
  title={Visual-tactile learning of garment unfolding for robot-assisted dressing},
  author={Zhang, Fan and Demiris, Yiannis},
  journal={IEEE Robotics and Automation Letters},
  year={2023},
  publisher={IEEE}
}

@article{lee2024gog,
  title={GOG: A Versatile Gripper-On-Gripper Design for Bimanual Cloth Manipulation with a Single Robotic Arm},
  author={Lee, Dongmyoung and Chen, Wei and Chen, Xiaoshuai and Rojas, Nicolas},
  journal={IEEE Robotics and Automation Letters},
  year={2024},
  publisher={IEEE}
}

@inproceedings{hietala2022learning,
  title={Learning visual feedback control for dynamic cloth folding},
  author={Hietala, Julius and Blanco--Mulero, David and Alcan, Gokhan and Kyrki, Ville},
  booktitle={2022 IEEE/RSJ International Conference on Intelligent Robots and Systems (IROS)},
  pages={1455--1462},
  year={2022},
  organization={IEEE}
}

@article{ebert2018visual,
  title={Visual foresight: Model-based deep reinforcement learning for vision-based robotic control},
  author={Ebert, Frederik and Finn, Chelsea and Dasari, Sudeep and Xie, Annie and Lee, Alex and Levine, Sergey},
  journal={arXiv preprint arXiv:1812.00568},
  year={2018}
}

@article{wu2019learning,
  title={Learning to manipulate deformable objects without demonstrations},
  author={Wu, Yilin and Yan, Wilson and Kurutach, Thanard and Pinto, Lerrel and Abbeel, Pieter},
  journal={arXiv preprint arXiv:1910.13439},
  year={2019}
}

@INPROCEEDINGS{10342086,
  author={Chen, Wei and Lee, Dongmyoung and Chappell, Digby and Rojas, Nicolas},
  booktitle={2023 IEEE/RSJ International Conference on Intelligent Robots and Systems (IROS)}, 
  title={Learning to Grasp Clothing Structural Regions for Garment Manipulation Tasks}, 
  year={2023},
  volume={},
  number={},
  pages={4889-4895},
  doi={10.1109/IROS55552.2023.10342086}}

@article{perez2017effectiveness,
  title={The effectiveness of data augmentation in image classification using deep learning},
  author={Perez, Luis and Wang, Jason},
  journal={arXiv preprint arXiv:1712.04621},
  year={2017}
}

@article{bjerrum2017smiles,
  title={SMILES enumeration as data augmentation for neural network modeling of molecules},
  author={Bjerrum, Esben Jannik},
  journal={arXiv preprint arXiv:1703.07076},
  year={2017}
}

@article{gulrajani2017improved,
  title={Improved training of wasserstein gans},
  author={Gulrajani, Ishaan and Ahmed, Faruk and Arjovsky, Martin and Dumoulin, Vincent and Courville, Aaron},
  journal={arXiv preprint arXiv:1704.00028},
  year={2017}
}

@article{azizi2023synthetic,
  title={Synthetic data from diffusion models improves imagenet classification},
  author={Azizi, Shekoofeh and Kornblith, Simon and Saharia, Chitwan and Norouzi, Mohammad and Fleet, David J},
  journal={arXiv preprint arXiv:2304.08466},
  year={2023}
}

@article{li2023synthetic,
  title={Is synthetic data from diffusion models ready for knowledge distillation?},
  author={Li, Zheng and Li, Yuxuan and Zhao, Penghai and Song, Renjie and Li, Xiang and Yang, Jian},
  journal={arXiv preprint arXiv:2305.12954},
  year={2023}
}

@Article{dhariwal2021diffusion,
  title={{Diffusion Models Beat GANs on Image Synthesis}},
  author={Dhariwal, Prafulla and Nichol, Alexander},
  journal={Advances in Neural Information Processing Systems},
  volume={34},
  pages={8780--8794},
  year={2021}
}

@article{ZHENG20201009,
title = {Conditional Wasserstein generative adversarial network-gradient penalty-based approach to alleviating imbalanced data classification},
journal = {Information Sciences},
volume = {512},
pages = {1009-1023},
year = {2020},
issn = {0020-0255},
doi = {https://doi.org/10.1016/j.ins.2019.10.014},
url = {https://www.sciencedirect.com/science/article/pii/S0020025519309715},
author = {Ming Zheng and Tong Li and Rui Zhu and Yahui Tang and Mingjing Tang and Leilei Lin and Zifei Ma}
}

@inproceedings{Karras2021,
  author = {Tero Karras and Miika Aittala and Samuli Laine and Erik H\"ark\"onen and Janne Hellsten and Jaakko Lehtinen and Timo Aila},
  title = {Alias-Free Generative Adversarial Networks},
  booktitle = {Proc. NeurIPS},
  year = {2021}
}

@inproceedings{
wang2023diffusiongan,
title={Diffusion-{GAN}: Training {GAN}s with Diffusion},
author={Zhendong Wang and Huangjie Zheng and Pengcheng He and Weizhu Chen and Mingyuan Zhou},
booktitle={The Eleventh International Conference on Learning Representations },
year={2023},
url={https://openreview.net/forum?id=HZf7UbpWHuA}
}

@inproceedings{
song2021denoising,
title={Denoising Diffusion Implicit Models},
author={Jiaming Song and Chenlin Meng and Stefano Ermon},
booktitle={International Conference on Learning Representations},
year={2021},
url={https://openreview.net/forum?id=St1giarCHLP}
}

@article{yuan2017gelsight,
  title = {{GelSight: High-Resolution Robot Tactile Sensors for Estimating Geometry and Force}},
  author={Yuan, Wenzhen and Dong, Siyuan and Adelson, Edward H.},
  journal={Sensors},
  volume={17},
  number={12},
  note={Art. no. E2762},
  year={2017}
}

@InProceedings{Kirillov_2023_ICCV,
    author    = {Kirillov, Alexander and Mintun, Eric and Ravi, Nikhila and Mao, Hanzi and Rolland, Chloe and Gustafson, Laura and Xiao, Tete and Whitehead, Spencer and Berg, Alexander C. and Lo, Wan-Yen and Dollar, Piotr and Girshick, Ross},
    title     = {{Segment Anything}},
    booktitle = {Proceedings of the IEEE/CVF International Conference on Computer Vision (ICCV)},
    month     = {October},
    year      = {2023},
    pages     = {4015-4026}
}

@INPROCEEDINGS{8460950,
  author={Marion, Pat and Florence, Peter R. and Manuelli, Lucas and Tedrake, Russ},
  booktitle={2018 IEEE International Conference on Robotics and Automation (ICRA)}, 
  title={Label Fusion: A Pipeline for Generating Ground Truth Labels for Real RGBD Data of Cluttered Scenes}, 
  year={2018},
  volume={},
  number={},
  pages={3235-3242},
  doi={10.1109/ICRA.2018.8460950}}

@INPROCEEDINGS{10532168,
  author={Lee, Dongmyoung and Chen, Wei and Rojas, Nicolas},
  booktitle={2024 10th International Conference on Mechatronics and Robotics Engineering (ICMRE)}, 
  title={Synthetic data enables faster annotation and robust segmentation for multi-object grasping in clutter}, 
  year={2024},
  volume={},
  number={},
  pages={253-259},
  doi={10.1109/ICMRE60776.2024.10532168}}

\end{document}